\documentclass{article}

    \PassOptionsToPackage{numbers, compress}{natbib}

\usepackage[main, final]{neurips_2026}

\usepackage[utf8]{inputenc} 
\usepackage[T1]{fontenc}    
\usepackage{hyperref}       
\usepackage{url}            
\usepackage{booktabs}       
\usepackage{amsfonts}       
\usepackage{nicefrac}       
\usepackage{microtype}      
\usepackage[dvipsnames,svgnames,x11names,hyperref]{xcolor}         
\usepackage{amsmath, amssymb, amsthm, mathtools}
\usepackage{algorithm}
\usepackage{algpseudocode}

\newtheorem{remark}{Remark}
\usepackage{thm-restate}

\usepackage{enumitem}
\usepackage{subcaption}
\usepackage{graphicx} 
\usepackage{multirow} 
\usepackage{booktabs}

\usepackage{soul}

\usepackage{thmtools}
\usepackage{thm-restate}
\usepackage{amsthm}
\usepackage[most]{tcolorbox}
\usepackage[capitalise,nameinlink]{cleveref}
\crefname{assumption}{Assumption}{Assumptions}

\hypersetup{
    colorlinks=true,
    citecolor=NavyBlue,
    linkcolor=NavyBlue,
    urlcolor=NavyBlue,
}
\title{Autoregressive Diffusion World Models for Off-Policy Evaluation of LLM Agents}

\author{%
  Kaixuan Liu \\
  Emory University \\
  \texttt{kaixuan.liu@emory.edu} \\
  \And
  Guojun Xiong\thanks{Correspondence to Guojun Xiong \texttt{<gjxiong@sjtu.edu.cn>}.} \\
  Shanghai Jiao Tong University \\
  \texttt{gjxiong@sjtu.edu.cn} \\
  \And
  Weinan Zhang \\
  Shanghai Jiao Tong University \\
  \texttt{wnzhang@sjtu.edu.cn} \\
  \And
  Shengpu Tang \\
  Emory University \\
  \texttt{shengpu.tang@emory.edu} \\
}

\begin{document}

\maketitle

\begin{abstract}
Evaluating large language model (LLM) agents in multi-turn interactive environments is expensive and risky, as it requires online environment interaction. We propose \textsc{Adwm} (Autoregressive Diffusion World Model), an evaluation framework that estimates the performance of a new LLM agent policy purely from pre-collected trajectories. The core idea is to learn a latent diffusion world model that simulates how the environment responds to the evaluation policy, without ever executing it in the real environment. Existing diffusion-based OPE methods guide full trajectories in a single pass by jointly diffusing states and actions, an assumption that breaks down for LLM agents whose actions are discrete text that must be sampled from the policy after observing the environment. Unlike textual world models that generate each observation token by token, \textsc{Adwm} jointly denoises the complete next observation in latent space, enabling step-by-step rollouts where the world model and agent alternate in causal order. Crucially, the LLM agent under evaluation directly guides the diffusion generation at each step via a policy-conditioned score function, ensuring that simulated trajectories accurately reflect its decision-making patterns. Empirically, \textsc{Adwm} achieves accurate policy rankings and evaluation reliability across diverse multi-turn agent tasks, demonstrating its promise as a practical framework for offline LLM agent evaluation. Code is available at \url{https://github.com/kaixuanliu-ai/ADWM}.
\end{abstract}

\section{Introduction}

Large language model agents are increasingly deployed in multi-turn
interactive environments, navigating websites, writing and executing
code, and reasoning over long document
contexts~\citep{yao2022react, zhou2023webarena, jimenez2023swe}.
As these agents are trusted with higher-stakes tasks, evaluating a
new agent policy before deployment becomes critical. Yet evaluation is expensive; every new agent must be executed live
in the real environment, consuming API budget and potentially
causing irreversible side effects.

Off-policy evaluation (OPE) addresses this challenge by estimating the value
of a new agent from pre-collected offline data, without any further
environment interaction. Importance sampling methods reweight offline
trajectories by policy likelihood ratios~\citep{liu2018breaking,
metelli2021subgaussian}, but these weights grow exponentially with
trajectory length, making the estimator impractical for multi-turn
agentic settings. Value-based direct methods attempt to fit a value function
on offline data~\citep{le2019batch,paine2020hyperparameter}, but suffer
from compounding bias when the evaluation policy differs substantially
from the behavior policy. Doubly robust methods combine the
two~\citep{farajtabar2018more, kallus2022efficiently}, partially
mitigating these issues but not resolving the fundamental difficulty
of long-horizon distribution shift in high-dimensional text spaces.

Among direct methods, a more promising model-based alternative is 
to build a world model: learn the environment dynamics from offline 
data and simulate rollouts under the evaluation policy~\citep{hanna2017bootstrapping, lu2023synthetic}.
In the agentic setting, this simulation must be \emph{autoregressive}: each LLM action depends on the observation just received, so the world model and agent must alternate step by step in causal order. Autoregressive transformer-based world models~\citep{micheli2022transformers} follow this structure naturally, but generate observations token by token, causing errors to accumulate within each step and compound further across steps, a critical failure mode for the long-horizon tasks that LLM agents face. Diffusion models offer a principled remedy: by jointly denoising the complete next observation, they remove the token-level error chain within each step~\citep{ho2020denoising, dhariwal2021diffusion}, and score-function guidance enables the evaluation policy to steer rollouts without retraining. However, existing diffusion-based methods for offline RL and OPE~\citep{jackson2024policy, lu2023synthetic, stitch} were designed for continuous control, where states and actions are both real-valued vectors that can be jointly noised and denoised as a single tensor in one pass, with actions generated by the diffusion process itself and the evaluation policy entering only as a guidance signal. In LLM agent settings, this assumption fundamentally breaks down: actions are discrete text that must be sampled from the LLM \emph{after} observing the environment and cannot participate in a joint continuous diffusion over trajectories. Generating full trajectories in a single pass therefore requires all LLM actions to be known before any observation is generated, a circular dependency that is irreconcilable with the step-by-step rollout that agentic evaluation demands.

\begin{figure*}[t]
\centering
\includegraphics[width=\textwidth]{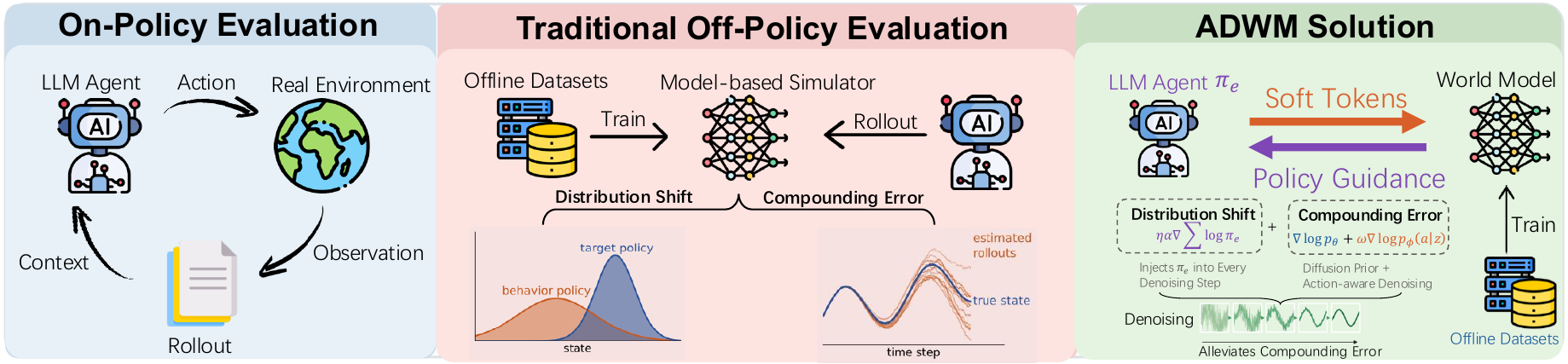}
\caption{\textbf{Comparison of evaluation paradigms for LLM agents.}
\textit{(Left)} On-policy evaluation requires executing the agent in
the real environment, which is expensive and potentially unsafe.
\textit{(Middle)} Traditional off-policy evaluation learns a model-based
simulator from offline data, but suffers from two fundamental issues:
distribution shift between the behavior and target policies, and
compounding error accumulated over multi-step rollouts.
\textit{(Right)} Our \textsc{Adwm} addresses both issues simultaneously.
Policy guidance injects $\pi_e$ into every denoising step to
alleviate distribution shift, while the diffusion prior combined with
action-aware denoising alleviates compounding error.}
\label{fig:teaser}
\end{figure*}

We propose \textbf{\textsc{Adwm}} (Autoregressive Diffusion World Model),
an offline evaluation framework for LLM agents that resolves this
tension by instantiating the world model itself as a diffusion
process (see Figure~\ref{fig:teaser}). The key insight is that the
globally policy-guided trajectory law can be \emph{exactly} factored
into a sequence of single-step conditionals, each decomposing as a
product of a prior, an action-posterior, and a policy continuation
factor (Theorem~\ref{thm:fused}). This product structure maps
directly onto guided diffusion, with the evaluation policy steering
every denoising step via its log-likelihood gradient. Because each
next observation is denoised jointly, \textsc{Adwm} avoids the
within-step token-level error chain, although transition errors may
still propagate across steps. \textsc{Adwm} is trained entirely offline,
and a new agent only requires fitting a lightweight adapter.

Our contributions are summarized as follows:\\
$\triangleright$ We derive the exact autoregressive single-step conditional
induced by a policy-guided trajectory law, showing it decomposes
into a principled three-factor product (Theorem~\ref{thm:fused}).\\
$\triangleright$ We realize this decomposition as a guided diffusion process
in a structured latent space, where the evaluation policy
participates in every denoising step without retraining the world
model.\\
$\triangleright$ We demonstrate empirically that \textsc{Adwm} correctly ranks evaluation
policies across diverse LLM agent benchmarks, outperforming
classic OPE baselines under
realistic LLM importance ratios.
\section{Related Work}

\paragraph{Model-Based OPE.}
Model-based OPE methods estimate policy value by learning a transition model from offline data and simulating rollouts under the evaluation policy~\citep{yu2020mopo, yu2021combo}. The core difficulty is compounding model error: small inaccuracies in the learned transition accumulate across steps, causing simulated trajectories to drift away from the true environment dynamics over long horizons~\citep{janner2019trust, sutton1996model, asadi2019combating}. Existing approaches mitigate this through pessimistic value penalties or conservative policy optimization~\citep{jin2021pessimism}, but these techniques were designed for low-dimensional continuous control and do not extend to free-form text observations. Beyond error accumulation, these methods provide no mechanism for conditioning the simulated rollout on the evaluation policy, so generated trajectories reflect the behavior distribution rather than what the evaluation policy would actually produce~\citep{voloshin2019empirical, feng2020accountable}. \textsc{Adwm} addresses both issues: jointly denoising each next observation removes the token-level error chain within a step~\citep{janner2022planning, ajay2022conditional}, while the guided score function ensures every generated observation is conditioned on the evaluation policy.

\paragraph{Diffusion Models for World Modeling.}
Diffusion models have emerged as powerful environment simulators that generate each transition through an iterative denoising process rather than token by token. Full-sequence models jointly generate entire trajectories~\citep{janner2022planning, ajay2022conditional, huang2025vid2world}, treating states and actions as a single jointly diffused tensor. This design precludes step-by-step rollout, as actions are produced by the diffusion process itself rather than sampled from an external policy. Score-based guidance~\citep{ho2022classifier} can steer such generation toward a target policy~\citep{lu2023synthetic, jackson2024policy}, but inherits the same limitation: actions must be real-valued vectors that participate in the joint diffusion, an assumption that breaks down for LLM agents whose actions are discrete text sampled from the policy after observing the environment. \textsc{Adwm} resolves this by deriving an exact factored score that enables the evaluation policy to steer each denoising step autoregressively, without requiring actions to be part of the diffusion process. Autoregressive video diffusion models such as Diffusion Forcing~\citep{chen2024diffusion} and Self Forcing~\citep{huang2025self} generate frames sequentially by assigning per-frame noise levels and by training on self-generated histories, respectively. These methods study sequential generation within a single diffusion model. \textsc{Adwm} instead alternates a diffusion world model with an external evaluation policy whose action is available only after it observes the current observation, so the interaction trajectory cannot be jointly diffused without violating causal order.

\paragraph{Textual World Models for LLM Agents.}
LLMs themselves can act as world models that predict environment feedback in text. SURGE~\citep{lyu2025surge} studies LLMs as surrogate code executors, CWM~\citep{copet2025cwm} trains a code LLM on execution traces and agentic interactions, and RAP~\citep{hao2023reasoning} uses an LLM world model for planning during reasoning. These simulators generate each observation token by token, so an early token error changes the prefix of all remaining tokens within the same environment step. Moreover, they target code execution, environment simulation, or online planning rather than OPE. \textsc{Adwm} learns from offline behavior trajectories, denoises the complete next-observation latent jointly, and estimates the value of a distinct evaluation policy without additional real-environment interaction.

\section{Problem Formulation}

\subsection{Agentic Reinforcement Learning}
\label{sec:agentic_rl}

We formulate agentic reinforcement learning as a sequential
decision-making problem, modeled as a Partially Observable
Markov Decision Process (POMDP) $\mathcal{M} = (\mathcal{S},
\mathcal{A}, \mathcal{O}, P, \mathcal{E}, r, \mu, \gamma)$,
where $\mathcal{S}$ is the state space, $\mathcal{A}$ is the
action space, $\mathcal{O}$ is the observation space,
$P: \mathcal{S} \times \mathcal{A} \to \Delta(\mathcal{S})$
is the transition kernel,
$\mathcal{E}: \mathcal{S} \to \Delta(\mathcal{O})$
is the emission distribution,
$r: \mathcal{S} \times \mathcal{A} \to \mathbb{R}$ is the
reward function, $\mu \in \Delta(\mathcal{S})$ is the initial
state distribution, and $\gamma \in [0,1)$ is the discount factor.
Since the agent does not have direct access to $s_t$, it instead
receives a partial observation $o_t \sim \mathcal{E}(\cdot \mid s_t)$
emitted by the environment. The agent interacts with the environment over $T$ turns. At each turn $t \in \{1, \dots, T\}$, the environment is in
state $s_t \in \mathcal{S}$, from which the agent receives
observation $o_t \in \mathcal{O}$ and selects an action
$a_t \sim \pi(\cdot \mid h_t)$ based on the interaction history:
\[
h_t = (o_1, a_1, \dots, o_{t-1}, a_{t-1}, o_t).
\]
The environment then transitions to the next state
$s_{t+1} \sim P(\cdot \mid s_t, a_t)$ and emits a reward
$r_t = r(s_t, a_t)$. A complete trajectory
$\tau = (o_1, a_1, r_1, \dots, o_T, a_T, r_T)$
has discounted return \mbox{$R(\tau) = \sum_{t=1}^{T} \gamma^{t-1} r_t$},
and induces the trajectory distribution:
\[
p_{\pi}(\tau) = \int \mu(s_1) \prod_{t=1}^{T}
\mathcal{E}(o_t \mid s_t)\,\pi(a_t \mid h_t)\, P(s_{t+1} \mid s_t, a_t)\, ds_{1:T+1}.
\]
The value of policy $\pi$ is thus
$J(\pi) = \mathbb{E}_{\tau \sim p_{\pi}}[R(\tau)]$. 
For example, in WebShop~\citep{yao2022webshop}, $o_t$ is the current web
page together with the task instruction, the LLM reads $h_t$ and outputs a
text action $a_t$ such as \texttt{search[...]} or
\texttt{click[buy now]}, and the environment returns the next page
$o_{t+1}$, the reward $r_t$, and a termination signal. A logged episode
annotated with this notation is given in \cref{app:webshop_example}.

\subsection{Off-Policy Evaluation for LLM Agents}
\label{sec:ope_formulation}

We assume access to an offline dataset:
\[
\mathcal{D} = \{\tau^{(i)}\}_{i=1}^{N},
\qquad \tau^{(i)} \sim p_{\pi_b}(\tau),
\]
collected by a behavior policy $\pi_b$, which may correspond
to a mixture of different LLM configurations and is
treated as unknown. Given $\mathcal{D}$ and a target evaluation policy $\pi_e$,
a language model distinct from $\pi_b$ that maps observation
history $h_t$ to a distribution over text actions, the goal
of off-policy evaluation (OPE) is to estimate:
\[
J(\pi_e) = \mathbb{E}_{\tau \sim p_{\pi_e}}[R(\tau)],
\]
without executing $\pi_e$ in the real environment.
The core challenge is distribution shift: since
$p_{\pi_b}(\tau) \neq p_{\pi_e}(\tau)$, trajectories
in $\mathcal{D}$ are not representative samples from
$p_{\pi_e}$, and direct Monte Carlo estimation on
$\mathcal{D}$ is severely biased. We address this challenge by learning a diffusion world model that simulates rollouts under $\pi_e$ directly, bypassing the need for explicit density ratio estimation.

\subsection{Denoising Diffusion Models}
\label{sec:ddpm}

Denoising Diffusion Probabilistic Models
(DDPMs)~\citep{ho2020denoising} learn to approximate a
target distribution $q(x)$ via two coupled Markov chains.
The forward process gradually corrupts a clean sample
$x^0 \sim q(x)$ over $K$ steps:
\[
x^k = \sqrt{\bar\alpha_k}\, x^0
+ \sqrt{1-\bar\alpha_k}\,\epsilon,
\qquad \epsilon \sim \mathcal{N}(0, I),
\]
where $\{\bar\alpha_k\}_{k=1}^{K}$ is a decreasing noise
schedule with $\bar\alpha_K \approx 0$. The reverse process
learns to recover $x^0$ from $x^K$ by iteratively applying
$p_\theta(x^{k-1} \mid x^k) = \mathcal{N}(x^{k-1};\,
\mu_\theta(x^k, k),\, \sigma_k^2 I)$,
where $\mu_\theta$ is parameterized via a noise prediction
network $\epsilon_\theta$ trained by:
\[
\mathcal{L}(\theta) =
\mathbb{E}_{k,\, x^0,\, \epsilon \sim \mathcal{N}(0,I)}
\bigl[\,\|\epsilon - \epsilon_\theta(x^k, k)\|^2\,\bigr].
\]

\paragraph{Guided diffusion.}
A trained diffusion model can be steered to sample from
a conditional distribution $p(x \mid y)$ without retraining,
by incorporating a guidance signal $y$ into each denoising
step~\citep{dhariwal2021diffusion}. Since the backward
transition is well approximated by a Gaussian,
$p(x^k \mid x^{k+1}) \approx \mathcal{N}(\mu_k, \Sigma_k)$,
Bayes' rule gives:
\[
p(x^k \mid x^{k+1}, y)
\propto
p(x^k \mid x^{k+1})\, p(y \mid x^k).
\]
Applying a first-order Taylor expansion of
$\log p(y \mid x^k)$ around $\mu_k$ yields:
\[
p(x^k \mid x^{k+1}, y)
\approx \mathcal{N}\!\bigl(
x^k;\; \mu_k + \Sigma_k g_k,\; \Sigma_k
\bigr),
\qquad
g_k = \nabla_{x} \log p(y \mid x)\big|_{x=\mu_k}.
\]
Sampling from the guided distribution therefore reduces
to shifting the denoising mean by $\Sigma_k g_k$ at each
step, with no modification to the model itself. In our
framework, $y$ is instantiated as the target LLM agent
$\pi_e$, and $g_k$ becomes the log-likelihood gradient
of $\pi_e$ with respect to the current latent state,
steering world-model rollouts toward trajectories
compatible with the agent under evaluation.
\section{Autoregressive Diffusion World Models}
\label{sec:method}

\begin{figure*}[!t]
\centering
\includegraphics[width=\textwidth]{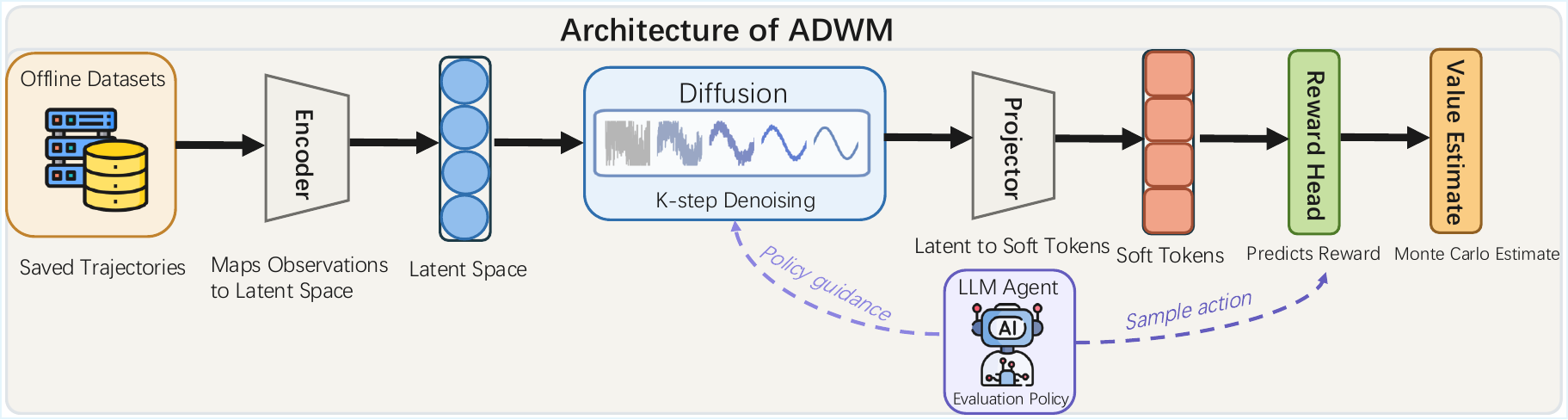}
\caption{\textbf{Architecture of \textsc{Adwm}.}
Offline trajectories are encoded by $E$ into latent states
$z_t$, which are processed by a diffusion world model
$p_\theta$ through $K$-step denoising. A projector $G_\psi$
maps each latent to soft tokens $\tilde{o}_t$ that the
evaluation policy $\pi_e$ can read in its own embedding
space. $\pi_e$ plays two complementary roles: it steers
the denoising process via policy guidance (dashed arrow,
Section~\ref{sec:guided_diffusion}), and samples actions
$a_t$ conditioned on $\tilde{o}_t$ to drive autoregressive
rollout (solid arrow). A reward head $r_\rho$ predicts
rewards $\hat{r}_t$ used in the Monte Carlo value estimate
$\hat{J}(\pi_e)$.}
\label{fig:architecture}
\end{figure*}

We develop an autoregressive diffusion world model for
off-policy evaluation of LLM agents. The core insight
is that a globally policy-guided trajectory law can be
exactly factored into a sequence of single-step
conditionals, each fusing local action consistency with
global policy guidance. We realize these conditionals
as a guided diffusion process in a structured latent
space, where the evaluation LLM agent directly steers the
denoising at every step (Figure~\ref{fig:architecture}).

\subsection{From Trajectory Guidance to Single-Step Conditionals}
\label{sec:ar_projection}

In our agentic setting, the behavior policy $\pi_b$ is
unknown and generally intractable, as trajectories in
$\mathcal{D}$ may originate from different LLM
configurations, prompting strategies, or agent versions.
A natural starting point is importance sampling, which
corrects for the mismatch between $p_{\pi_b}$ and
$p_{\pi_e}$ via policy-likelihood ratios. Since the
environment transition cancels between $p_{\pi_e}$ and
$p_{\pi_b}$, the target trajectory law satisfies:
\begin{align}
p_{\pi_e}(\tau)
= p_{\pi_b}(\tau)\cdot
\prod_{t=1}^{T}
\frac{\pi_e(a_t \mid h_t)}{\pi_b(a_t \mid h_t)},
\label{eq:is_identity}
\end{align}
so $\log p_{\pi_e}(\tau) = \log p_{\pi_b}(\tau)
+ \sum_t \log \pi_e(a_t \mid h_t)
- \sum_t \log \pi_b(a_t \mid h_t)$.
Since $\pi_b$ is unavailable, and dropping it acts as
a form of behavior regularization~\citep{jackson2024policy}
that anchors the distribution to the support of
$\mathcal{D}$, we learn a generative model $p_\theta(\tau)$
of the logged trajectory distribution via maximum
likelihood, and introduce $\alpha > 0$ to control the
correction strength, yielding the policy-guided target law:
\begin{equation}
q_{\alpha}(\tau\mid\pi_{e})
\;\propto\;
p_{\theta}(\tau)\,
\exp\!\Big(\alpha\sum_{t=1}^{T}\log\pi_{e}(a_{t}\mid h_{t})\Big),
\qquad \alpha>0.
\label{eq:pgd_target}
\end{equation}
When $p_\theta = p_{\pi_b}$ and $\alpha = 1$, this
recovers $p_{\pi_e}$ exactly; the full derivation is
in Appendix~\ref{app:is_derivation}.

The target law \eqref{eq:pgd_target} is defined over full trajectories, but cannot be instantiated directly for a black-box $\pi_e$. Since $\pi_e$ is a black-box LLM that must observe $o_t$ before producing $a_t$, the trajectory-level tilt $\sum_t \log \pi_e(a_t \mid h_t)$ cannot be evaluated without first generating all observations in causal order. This forces an autoregressive decomposition: the world model and $\pi_e$ must alternate step by step, generating observations and actions one at a time. The following proposition, proved in Appendix~\ref{app:proof}, characterizes the exact single-step conditional induced by \eqref{eq:pgd_target}.

\begin{restatable}[Autoregressive conditional law]{proposition}{ARConditionalLaw}
\label{prop:ar_projection}
Let $q_{\alpha}$ be the policy-guided trajectory law
\eqref{eq:pgd_target}. Then for every $(h_{t},a_{t})$:
\begin{equation}
q_{\alpha}(o_{t+1}\mid h_{t},a_{t};\pi_{e})
\;\propto\;
p_{\theta}(o_{t+1}\mid h_{t},a_{t})\,
\mathcal{C}_{\alpha}(o_{t+1};h_{t},a_{t},\pi_{e}),
\label{eq:ar_projection}
\end{equation}
where the \emph{continuation factor} is:
\begin{equation}
\mathcal{C}_{\alpha}(o_{t+1};h_{t},a_{t},\pi_{e})
\;:=\;
\mathbb{E}_{\tau_{t+1:T}\sim p_{\theta}(\cdot\mid h_{t},a_{t},o_{t+1})}
\!\left[
\exp\!\Big(
\alpha\sum_{u=t+1}^{T}\log\pi_{e}(a_{u}\mid h_{u})
\Big)
\right].
\label{eq:continuation}
\end{equation}
\end{restatable}

Two candidates for $o_{t+1}$ may be equally supported by
$p_\theta(o_{t+1} \mid h_t, a_t)$ yet lead to futures
with very different compatibility with $\pi_e$;
$\mathcal{C}_\alpha$ is precisely the term that
distinguishes them. A Bayesian decomposition of
$p_\theta(o_{t+1} \mid h_t, a_t)$ further separates the
prior dynamics from the influence of $a_t$, giving the
central result of our framework.

\begin{restatable}[Fused single-step conditional]{theorem}{FusedConditional}
\label{thm:fused}
Let $q_{\alpha}$ be the policy-guided trajectory law
\eqref{eq:pgd_target}. For every $(h_t, a_t)$ and any
$\omega > 0$, $\eta \geq 0$:
\begin{equation}
\widetilde{q}_{\omega,\eta}(o_{t+1}\mid h_{t},a_{t};\pi_{e})
\;\propto\;
\underbrace{p_{\theta}(o_{t+1}\mid h_{t})}_{\text{prior}}
\;\cdot\;
\underbrace{p_{\theta}(a_{t}\mid o_{t+1},h_{t})^{\omega}}_{\text{action posterior}}
\;\cdot\;
\underbrace{\mathcal{C}_{\alpha}(o_{t+1};h_{t},a_{t},\pi_{e})^{\eta}}_{\text{policy continuation}},
\label{eq:fused}
\end{equation}
where $\omega=\eta=1$ recovers the exact conditional
induced by $q_\alpha$.
\end{restatable}

\begin{remark}[Limiting cases of $\omega$ and $\eta$]
The parameters $\omega$ and $\eta$ interpolate between
degenerate regimes that illuminate the role of each
guidance term. Setting $\eta = 0$ removes the policy
continuation factor entirely, reducing
$\widetilde{q}_{\omega,0}$ to an action-posterior-weighted
prior that simulates the behavior distribution with no
correction toward $\pi_e$. Setting $\omega = 0$ suppresses
the action posterior, causing the model to generate
observations from the unconditional prior
$p_\theta(o_{t+1}\mid h_t)$ irrespective of $a_t$ and
breaking action-consistency. The canonical choice
$\omega=\eta=1$ recovers the exact conditional induced
by $q_\alpha$. In practice, $\omega$ is controlled by
the CFG scale $\lambda$ via $\omega=1+\lambda$, and
$\eta$ is annealed over the reverse process; both are
treated as hyperparameters rather than tuned per
environment.
\end{remark}

The three factors play distinct roles. The prior $p_\theta(o_{t+1} \mid h_t)$ anchors generated observations to the support of the offline data. The action posterior $p_\theta(a_t \mid o_{t+1}, h_t)^\omega$ injects current-step action information without a separate classifier, recovered directly from classifier-free guidance as shown in Section~\ref{sec:guided_diffusion}. The continuation factor $\mathcal{C}_\alpha$ re-weights candidate observations by their long-horizon compatibility with $\pi_e$, ensuring the world model generates observations that open up futures where $\pi_e$ can act effectively. Crucially, since \eqref{eq:fused} is a single-step conditional, $\pi_e$ never needs to produce actions before observations are generated, fully resolving the circular dependency that prevents existing diffusion OPE methods from operating autoregressively. The proof is given in Appendix~\ref{app:fusion}. Each factor corresponds to a distinct gradient term in the score function, which we now realize as a guided diffusion process in latent space.

Since \eqref{eq:fused} is a product of three factors, its log decomposes
into a sum, and differentiating with respect to $o_{t+1}$ yields an
additive score:
\begin{equation}
\nabla_{o_{t+1}}\log\widetilde{q}_{\omega,\eta}
\;=\;
\underbrace{\nabla\log p_\theta(o_{t+1}\mid h_t)}_{\text{prior score}}
\;+\;
\omega\underbrace{\nabla\log p_\theta(a_t\mid o_{t+1},h_t)}_{\text{action-posterior score}}
\;+\;
\eta\underbrace{\nabla\log\mathcal{C}_\alpha(o_{t+1};h_t,a_t,\pi_e)}_{\text{continuation score}}.
\label{eq:score_decomp}
\end{equation}
This additive structure is precisely what makes guided diffusion
applicable: each term can be approximated and injected independently
into the reverse process without retraining the world model. The prior
score is given directly by the score network; the action-posterior score
is recovered from the difference between action-conditioned and
unconditional outputs via classifier-free guidance; and the continuation
score requires differentiating through $\pi_e$'s log-likelihoods along
future rollouts. We realize all three in a structured latent space, as
detailed next.

\subsection{Guided Diffusion World Model}
\label{sec:guided_diffusion}

The product structure of Theorem~\ref{thm:fused} maps directly
onto guided diffusion: each factor contributes an additive term
to the score, and sampling reduces to running a guided reverse
process with $\pi_e$ steering the denoising at every step.
To instantiate this in practice, we first specify the latent
space in which the diffusion operates and how $\pi_e$ reads
world-model states.

\paragraph{Latent state and policy interface.}
Observations in LLM agent tasks are raw text, which cannot be
directly denoised. We therefore operate in a structured latent
space. An encoder $E$ maps each observation to a compact
representation $z_t = E(o_t)\in\mathbb{R}^d$, and a projector
$G_\psi$ produces soft tokens
$\tilde{o}_t = G_\psi(z_t)\in\mathbb{R}^{n_{\mathrm{soft}}\times d_{\mathrm{LLM}}}$
that $\pi_e$ can read in its own embedding space. The world model
operates on the latent history $h_t=(z_1,a_1,\dots,z_t)$, while
$\pi_e$ conditions on $h_t^\pi=(\tilde{o}_1,a_1,\dots,\tilde{o}_t)$.
A pretrained semantic encoder is insufficient for this role: two
observations that differ only in a negation may be nearby in
semantic space yet lead to entirely different dynamics, corrupting
all three guidance terms in \eqref{eq:final_score} simultaneously.
We therefore train $E$ end-to-end jointly with the world-model
components on $\mathcal{D}$, so that the representation is shaped
by the dynamics of the environment rather than surface semantics.
The projector $G_\psi$ is then fitted to the evaluation LLM with a
contrastive InfoNCE objective~\citep{oord2018representation}:
\begin{equation}
\mathcal{L}_{\psi} = -\frac{1}{2B}\sum_{i=1}^{B}
\left[
\log\frac{e^{s_{ii}/\kappa}}{\sum_{j}e^{s_{ij}/\kappa}}
+
\log\frac{e^{s_{ii}/\kappa}}{\sum_{j}e^{s_{ji}/\kappa}}
\right]
+
\frac{\beta}{B}\sum_{i=1}^{B}
\bigl\|\bar{G}_\psi(z_i) - \bar{e}_i\bigr\|_2^2,
\label{eq:infonce}
\end{equation}
where $s_{ij} = \mathrm{cossim}(\bar{G}_\psi(z_i), \bar{e}_j)$,
$\bar{G}_\psi(z_i)$ is the token-mean of the projected soft
tokens, $\bar{e}_i$ is the token-mean of the ground-truth
$\pi_e$ input embeddings, $\kappa$ is a temperature, and
$\beta=0.1$ balances discriminability against distributional
alignment with $\pi_e$'s embedding space. Two auxiliary losses shape the representation: an inverse
dynamics loss $\mathcal{L}_{\mathrm{IDM}}$ encouraging action-aware
latents, and a behavior cloning loss $\mathcal{L}_{\mathrm{BC}}$
providing a stable retrieval target for $G_\psi$; full definitions
are in \cref{app:wm-loss}.

\paragraph{Guided score.}
Following classifier-free guidance~\citep{ho2022classifier}, we
parameterize the world model as a score network
$s_\theta(z_{t+1}^k, k; h_t^z, a_t)$. A causal Transformer encodes the
latent history $h_t^z$ from the sequence of $(z_u, a_u)$ pairs with a
subsequent-step mask, so the summary at step $t$ depends only on past
latents and actions. The score network is an MLP that receives the noised
latent $z_{t+1}^k$, the diffusion step $k$, this history summary, and the
action embedding through FiLM conditioning (\cref{app:wm-arch}). To realize the prior
and action-posterior factors of Theorem~\ref{thm:fused} without
a classifier, we train $s_\theta$ under two regimes by
randomly masking $a_t$ with probability $p$ during training:
\begin{align}
\mathcal{L}_{\mathrm{DSM}}(\theta)
=
\mathbb{E}\Big[
\big\|s_\theta(z_{t+1}^k, k; h_t^z, \tilde{a}_t)
- \nabla_{z^k}\log p_k(z_{t+1}^k \mid z_{t+1})\big\|^2
\Big],
\quad
\tilde{a}_t =
\begin{cases}
\emptyset & \text{w.p. } p,\\
a_t & \text{otherwise.}
\end{cases}
\label{eq:dsm_loss}
\end{align}
By Bayes' rule, the difference between the action-conditioned
and unconditional outputs recovers the score of the implicit action
posterior $p_\phi(a_t \mid z_{t+1}^k, h_t^z)$, so
classifier-free guidance with scale $\lambda$ directly realizes
the first two terms of~\eqref{eq:fused}, with $\omega = 1+\lambda$.

For the continuation term, the score
$\nabla_{z_{t+1}}\log\mathcal{C}_\alpha$ requires an expectation
over all future trajectories and is intractable. At each diffusion
step $k$, the score network provides a denoised estimate
$\hat{z}_{t+1}$; we roll the world model forward from this
estimate and differentiate through $\pi_e$'s log-likelihoods
along the resulting trajectory. The complete guided score is:
\begin{align}
s(z_{t+1}^{k},k;h_t^z,a_t,\pi_e)
&=
\nabla_{z_{t+1}^{k}}
\log p_{\theta}(z_{t+1}^{k}\mid h_t^z)
+
\omega\,
\nabla_{z_{t+1}^{k}}
\log p_{\phi}(a_t\mid z_{t+1}^{k},h_t^z)
\nonumber\\
&\quad
+\;\eta_k\,\alpha\,
\nabla_{\hat{z}_{t+1}}
\sum_{u=t+1}^{T}
\log \pi_e(\hat{a}_u\mid h_u^\pi),
\label{eq:final_score}
\end{align}
where $\eta_k$ is annealed so that continuation guidance
strengthens as $\hat{z}_{t+1}$ becomes more reliable late
in the reverse process. The complete derivation of
\eqref{eq:final_score} from Theorem~\ref{thm:fused} is given
in Appendix~\ref{app:score_derivation}.

\paragraph{Value estimation.}
Given the guided transition model, we estimate $J(\pi_e)$ by
Monte Carlo rollout. At each step, $\pi_e$ samples an action
$a_t \sim \pi_e(\cdot \mid h_t^\pi)$, the world model generates
the next latent $z_{t+1}$ via the reverse diffusion process with
score \eqref{eq:final_score}, and the reward head predicts
$\hat{r}_t = r_\rho(z_{t+1}, h_t^z)$. Episodes
terminate when the termination head
$\hat{d}_t = \sigma(d_\rho(z_{t+1}, h_t^z))$ exceeds a threshold.
Averaging over $M$ rollouts gives:
\begin{align}
\widehat{J}(\pi_e)
= \frac{1}{M}\sum_{m=1}^{M}\sum_{t=1}^{T_m}
\hat{r}_t^{(m)},
\label{eq:mc_estimator}
\end{align}
where $T_m$ is the termination step of the $m$-th rollout.
The rollout follows the causal order of the real interaction: $\pi_e$
reads only the soft tokens $\tilde{o}_{1:t}$ produced by the world model,
and each next latent is generated only after $a_t$ is sampled. Hence no
environment interaction is required at evaluation time.
\section{Experiments}
\label{sec:experiments}

We evaluate \textsc{Adwm} around three questions:
\begin{enumerate}
    \item \emph{Can a world model rank unseen policies from behavior data alone?}
    \item \emph{How does it compare to classical OPE under realistic LLM importance ratios?}
    \item \emph{Which components of the diffusion guidance matter?}
\end{enumerate}

\paragraph{Experiment setup.}
\label{sec:setup}

We evaluate \textsc{Adwm} on four LLM-agent benchmarks chosen to span the reward
and post-training regimes of modern LLM agents. On the reward axis, the
suite covers dense per-step reward (HotpotQA F1~\citep{yang2018hotpotqa}),
shaped partial reward (ScienceWorld~\citep{wang2022scienceworld}),
continuous partial reward (WebShop~\citep{yao2022webshop}), and sparse
binary success (ALFWorld~\citep{shridhar2020alfworld}). On the policy
axis, the cells cover task-specific RLHF-style post-training (DPO and
PRM on HotpotQA~\citep{xiong2025rag}; ETO on
ScienceWorld~\citep{song2024trial}) and generic-to-specialized iterative
fine-tuning (LEAP on ALFWorld and
WebShop~\citep{choudhury2024better}). All evaluation policies are
publicly released checkpoints from prior work, summarised alongside
their behavior counterparts in \cref{tab:setup}. Representative logged interaction traces for WebShop and ALFWorld are
provided in \cref{app:webshop_example,app:alfworld_projection_example}.

\begin{table}[h]
\centering
\small
\setlength{\tabcolsep}{5pt}
\renewcommand{\arraystretch}{1.15}
\caption{Environments, policies, and reward structure. The behavior
LLM $\pi_b$ is used only to collect world-model training data; the
evaluation LLM $\pi_e$ is the policy \textsc{Adwm} scores at test time.
Citations point to the benchmark source and the released $\pi_e$
checkpoint. Per-cell behavior trajectory counts, GT episode budgets,
and full hyperparameters are in \cref{app:setup}.}
\label{tab:setup}
\begin{tabular}{@{}llll@{}}
\toprule
Env. & Behavior $\pi_b$ & Evaluation $\pi_e$ & Reward \\
\midrule
HotpotQA \citep{yang2018hotpotqa}
  & ReAct-HotpotQA-SFT
  & ReAct-HotpotQA \{DPO, PRM\} \citep{xiong2025rag}
  & dense F1 \\
ScienceWorld \citep{wang2022scienceworld}
  & sw-llama-sft
  & sw-llama-eto \citep{song2024trial}
  & partial \\
ALFWorld \citep{shridhar2020alfworld}
  & Llama-3.1-8B-Instr.
  & leap-alf-\{iter1, iter3\} \citep{choudhury2024better}
  & sparse 0/1 \\
WebShop \citep{yao2022webshop}
  & Llama-3.1-8B-Instr.
  & leap-ws-iter1 \citep{choudhury2024better}
  & cont.\ partial \\
\bottomrule
\end{tabular}
\end{table}

In every cell, the behavior policy $\pi_b$ used to collect world-model
training trajectories is strictly distinct from the evaluation policy
$\pi_e$ being scored. Concretely, $\pi_b$ is either the matching
task-specific SFT base of the same agent (HotpotQA, ScienceWorld) or,
when no such base has been publicly released, the generic
instruction-tuned Llama-3.1-8B-Instruct \citep{grattafiori2024llama}
(ALFWorld, WebShop); the ALFWorld-iter3 and WebShop pools also include two
other LEAP iterates (\cref{tab:behdata}). The world model is therefore
trained without ever observing $\pi_e$. On ALFWorld we additionally evaluate the third LEAP
iteration to test cross-iteration ranking; on HotpotQA we pool DPO and
PRM checkpoints into a 10-cell cross-policy split that probes
policy-level discrimination beyond the within-policy $\varepsilon$-axis.

To assess sensitivity across a graded range of policy quality, we
construct a one-parameter family of evaluation policies via
$\varepsilon$-greedy mixing: at each step, $\pi_e$ acts with probability
$1{-}\varepsilon$ and a uniformly random admissible action is taken with
probability $\varepsilon$, for $\varepsilon\in\{0,0.25,0.5,0.75,1.0\}$.
The real-environment ground-truth curve is obtained by executing
$\pi_e^\varepsilon$ in the actual benchmark; \textsc{Adwm} produces $\hat{J}$
entirely from world-model rollouts with no environment access at
evaluation time. We run five random seeds per cell and report the
Spearman rank correlation between the seed-averaged $\hat{J}$ curve and
the ground-truth curve. Full experimental details are in \cref{app:setup}.

The behavior and evaluation policies are not close. We measure the shift by
the cumulative trajectory log-likelihood ratio:
\[
\log\frac{p_{\pi_e}(\tau)}{p_{\pi_b}(\tau)}
=\sum_{t=1}^{T}\log\frac{\pi_e(a_t\mid h_t)}{\pi_b(a_t\mid h_t)}.
\]
At $\varepsilon{=}0.5$, its magnitude reaches up to $24$, $18$, $35$, and
$28$ nats on HotpotQA, ScienceWorld, ALFWorld, and WebShop, so the trajectory
likelihood ratios span roughly $16$ to $30$ orders of magnitude
(\cref{app:ratio_variance}). The evaluation checkpoint is also excluded from every
behavior pool. The experiments therefore go beyond the near-policy regime,
although they do not cover fully out-of-support or cross-family evaluation.

\paragraph{Baselines.}
We compare \textsc{Adwm} against five classical OPE estimators: direct method
(DM)~\citep{voloshin2019empirical}, importance sampling
(IS)~\citep{precup2000eligibility}, weighted importance sampling
(WIS)~\citep{precup2000eligibility}, fitted-Q evaluation
(FQE)~\citep{le2019batch}, and doubly robust estimation
(DR)~\citep{jiang2016doubly}, all implemented following the COBS
reference protocol~\citep{voloshin2019empirical}. For estimators that
require $\pi_b$ (IS, WIS, DR), we supply token log-probabilities of
the logged actions under the behavior LLM rather than the uniform action
prior commonly used in prior work, providing these baselines with
privileged information that is unavailable in realistic deployments
(closed-source APIs, deleted checkpoints, hidden tokenizers). We
include them as diagnostic references rather than competitive baselines,
since the unbounded LLM action space and limited distributional overlap
in long-horizon rollouts violate the absolute-continuity assumption
underlying IS-based estimators. Implementation details and the
per-$\varepsilon$ FQE correction are in \cref{app:baselines}.

\subsection{Results}
\label{sec:results}

\paragraph{Main results.}
\Cref{tab:main_results} compares \textsc{Adwm} against all five classical
estimators across six $(\pi_b,\pi_e)$ configurations. \textsc{Adwm} is the only
estimator with strictly positive Spearman $\rho$ in every cell, ranging
from $+0.67$ on ALFWorld-iter1 to $+0.90$ on both HotpotQA-DPO and
WebShop-iter1 (mean $+0.82$), while every classical baseline is non-positive in
at least two cells. IS and DR collapse under LLM-induced importance
ratios spanning up to thirty orders of magnitude, with correlations as
low as $-0.90$. WIS eliminates the explosion but degenerates to a
single dominant-weight trajectory over long horizons, producing
$\rho{=}0$ on both ALFWorld cells and spurious positive correlations
elsewhere driven by floating-point underflow ties. DM estimates a
behavior-policy state value invariant to the evaluation policy's action
distribution, yielding $\rho{=}0$ within every $\varepsilon$-sweep and
inverting on the cross-policy split ($\rho{=}{-}0.45$). FQE is the
sole partial exception: once corrected for the $\varepsilon$-linearity
artifact in the COBS protocol, it attains $\rho{=}{+}0.82$ on
ALFWorld-iter1 but collapses or inverts elsewhere (mean
$\rho{=}{+}0.10$). \textsc{Adwm}, requiring neither importance weights nor a
linear inductive bias, outperforms the next-best baseline by $+0.34$
in mean correlation; the world model is trained exclusively on $\pi_b$
trajectories in every cell. Per-seed statistics and $\hat{J}$/GT curves
are in \cref{app:results}.

\begin{table}[h]
\centering
\small
\setlength{\tabcolsep}{4pt}
\renewcommand{\arraystretch}{1.15}
\caption{Per-cell Spearman $\rho$ between each estimator's $\hat{J}$
and the real-environment ground-truth curve. \textsc{Adwm} is the only
estimator with $\rho{>}0$ on every configuration (mean $+0.82$, min
$+0.67$). CI$_{95}$ and $p$-values from 2000 bootstrap resamples
on the $(\hat J,\text{GT})$ pairs. Per-seed values and the underlying
$\hat J$ / GT curves are in \cref{app:results}.}
\label{tab:main_results}
\begin{tabular}{@{}lc ccc ccccc@{}}
\toprule
& & \multicolumn{3}{c}{\textsc{Adwm} (ours)} &
  \multicolumn{5}{c}{Classical OPE baselines (Spearman $\rho$)} \\
\cmidrule(lr){3-5}\cmidrule(lr){6-10}
Configuration & $n$ & $\rho$ & CI$_{95}$ & $p$ & FQE & DR & DM & IS & WIS \\
\midrule
HotpotQA-DPO     &  5 & $\mathbf{+0.90}$ & $[+0.11,+1.00]$ & $0.037$ & $-0.10$ & $-0.90$ & $\phantom{+}0.00$ & $-0.90$ & $\mathbf{+0.90}$ \\
ScienceWorld-ETO &  5 & $+0.82$ & $[+0.00,+1.00]$ & $0.089$ & $-0.21$ & $+0.21$ & $\phantom{+}0.00$ & $+0.21$ & $\mathbf{+0.97}$ \\
ALFWorld-iter1   &  5 & $+0.67$ & $[-0.91,+1.00]$ & $0.219$ & $\mathbf{+0.82}$ & $+0.21$ & $\phantom{+}0.00$ & $-0.82$ & $\phantom{+}0.00$ \\
ALFWorld-iter3   &  5 & $\mathbf{+0.80}$ & $[+0.11,+1.00]$ & $0.104$ & $\phantom{+}0.00$ & $+0.40$ & $\phantom{+}0.00$ & $-0.70$ & $\phantom{+}0.00$ \\
WebShop-iter1    &  5 & $\mathbf{+0.90}$ & $[+0.11,+1.00]$ & $0.037$ & $\phantom{+}0.00$ & $-0.20$ & $\phantom{+}0.00$ & $-0.90$ & $\mathbf{+0.90}$ \\
HotpotQA-cross   & 10 & $\mathbf{+0.81}$ & $[+0.23,+1.00]$ & $0.005$ & $+0.12$ & $-0.64$ & $-0.45$           & $-0.52$ & $+0.13$ \\
\midrule
Mean $\rho$      &    & $\mathbf{+0.82}$ & & & $+0.10$ & $-0.15$ & $-0.07$           & $-0.61$ & $+0.48$ \\
Min $\rho$       &    & $\mathbf{+0.67}$ & & & $-0.21$ & $-0.90$ & $-0.45$           & $-0.90$ & $\phantom{+}0.00$ \\
\bottomrule
\end{tabular}
\end{table}

\begin{figure}[h]
\centering
\includegraphics[width=\linewidth]{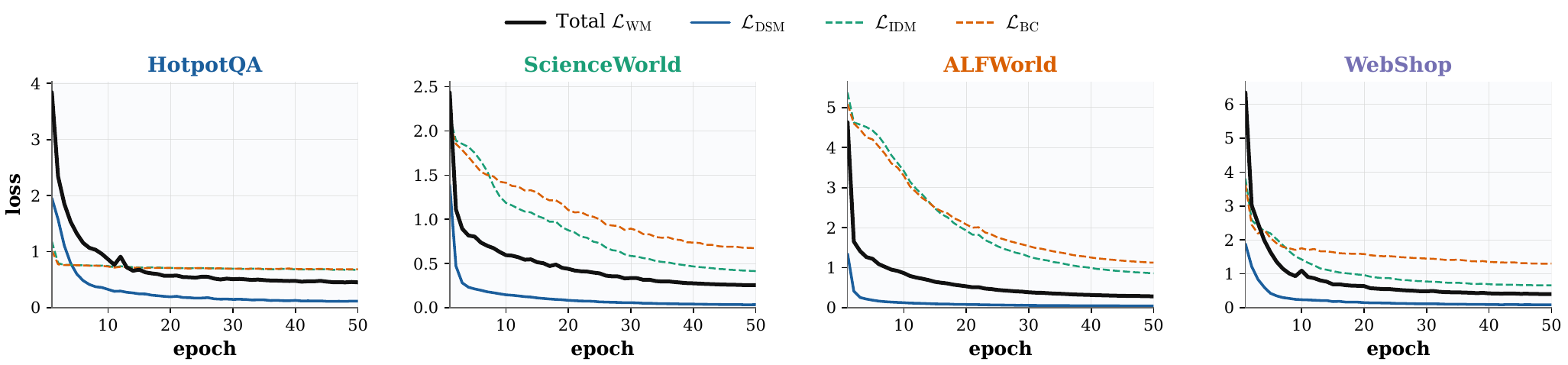}
\caption{Training loss curves across all four benchmarks. Total
world-model loss $\mathcal{L}_{\mathrm{WM}}$ and its components
converge stably within 50 epochs; see \cref{app:wm-loss} for full
definitions and \cref{app:wm-train} for the training procedure.}
\label{fig:training}
\end{figure}

\paragraph{Training convergence.}
\Cref{fig:training} shows that all loss components converge stably
across all four benchmarks within 50 epochs. Beyond
$\mathcal{L}_{\mathrm{DSM}}$ \eqref{eq:dsm_loss}, the total objective
$\mathcal{L}_{\mathrm{WM}}$ includes two auxiliary terms: an inverse
dynamics loss $\mathcal{L}_{\mathrm{IDM}}$ that predicts the bridging
action between consecutive latents, encouraging action-aware
representations, and a behavior cloning loss $\mathcal{L}_{\mathrm{BC}}$
that provides a stable retrieval target for the $\psi$ adapter.
$\mathcal{L}_{\mathrm{DSM}}$ converges fastest; $\mathcal{L}_{\mathrm{IDM}}$
and $\mathcal{L}_{\mathrm{BC}}$ converge more gradually, consistent
with the greater difficulty of predicting actions and cloning behavior
on long-horizon text observations. Full definitions and loss weights
are in \cref{app:wm-loss}.

\begin{figure}[h]
\centering
\includegraphics[width=0.92\linewidth]{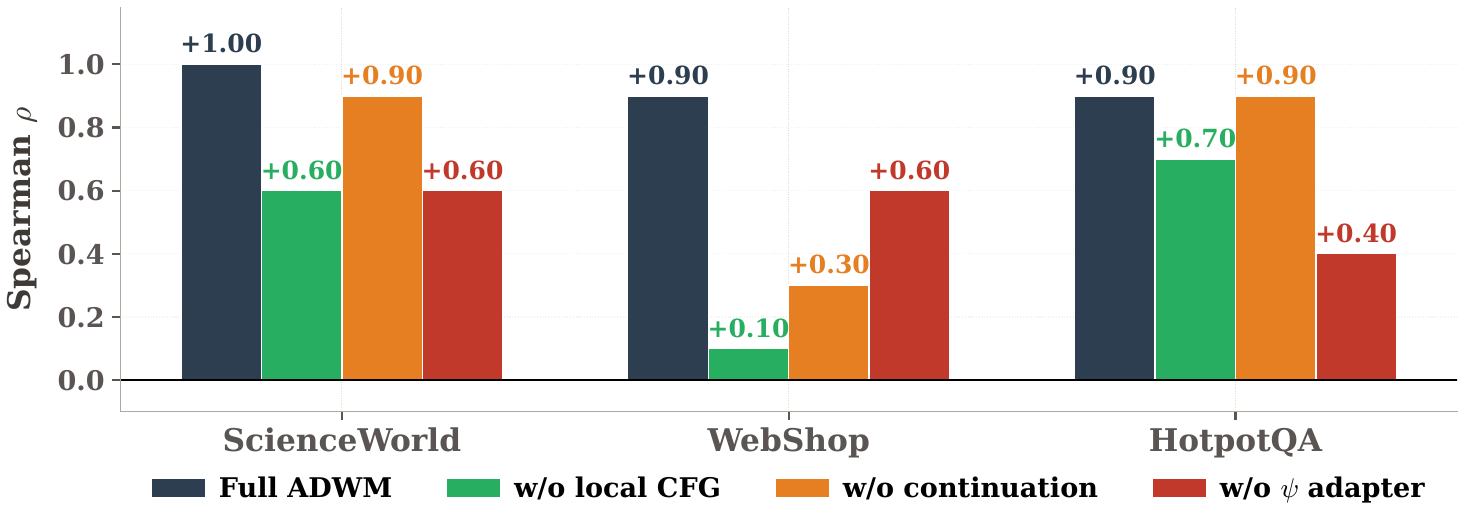}
\caption{Component ablation on three environments (avg-$\hat{J}$
Spearman $\rho$, single seed; five-seed trends in \cref{app:ablations}).
Each component dominates in a distinct reward regime: local CFG on
long-horizon partial-reward tasks, continuation guidance on goal-conditioned
trajectories, and the $\psi$ adapter on linguistically rich state
spaces. Removing any single component degrades at least one
environment; full \textsc{Adwm} consistently achieves the highest $\rho$.}
\label{fig:ablation}
\end{figure}

\paragraph{Ablations.}
\Cref{fig:ablation} isolates the contribution of \textsc{Adwm}'s three core
components across three environments with distinct reward structures.
Each component dominates in a different reward regime, and no single
component suffices across the full suite. Local CFG is most critical
on long-horizon partial-reward tasks: removing it collapses WebShop
$\rho$ from $+0.90$ to $+0.10$ and drops ScienceWorld from $+1.00$ to
$+0.60$, with a smaller drop on HotpotQA ($+0.90\to+0.70$). Continuation guidance
is the primary mechanism for keeping trajectories on a successful path
when per-step reward is absent, with removal degrading WebShop to
$+0.30$ and ScienceWorld to $+0.90$. The $\psi$ adapter dominates on
linguistically rich environments: its removal drops HotpotQA to $+0.40$
and degrades both ScienceWorld and WebShop to $+0.60$. \textsc{Adwm}'s
robustness therefore requires the combination of all three components.
Additional ablations on the latent dimension and on the design of the
$\psi$ adapter loss are given in \cref{app:ablations}.

\section{Limitations}
\label{sec:limitations_main}
The policy-ranking experiments rely mainly on $\varepsilon$-greedy variants of
a small number of publicly released base policies, because released agent
checkpoints with different performance levels on the same benchmark remain
scarce. Spearman $\rho$ measures ranking rather than the calibration of
absolute values, and we do not ablate the size or the behavior-policy diversity of the
offline data. \textsc{Adwm} removes the
token-level error chain within each observation, but transition errors may
still propagate across environment steps, and we do not provide an
end-to-end error bound. Finally, the soft-token projector is fitted to each evaluation LLM,
although neither the offline trajectories nor the transition model need
to be recollected or retrained.
A natural extension is a hybrid protocol that uses \textsc{Adwm} for offline
ranking and spends a small online budget only on the highest-ranked or
hardest-to-distinguish policies. Details are in
\cref{sec:limitations}.

\section{Conclusion}
We presented \textsc{Adwm}, an offline evaluation framework that estimates LLM
agent policy value purely from pre-collected trajectories. The key
contribution is an exact decomposition of a policy-guided trajectory
law into autoregressive single-step conditionals, each realized as a
guided diffusion process in which the evaluation policy steers every
denoising step without retraining or explicit importance weights.
Empirically, \textsc{Adwm} is the only estimator with consistently positive rank
correlations across all six $(\pi_b,\pi_e)$ configurations, with
ablations confirming that robustness requires the combination of all
three guidance components.

\begin{ack}
This work was supported in part by the National Natural Science Foundation of China (62322603) and the Shanghai Municipal Science and Technology Major Project (2025SHZDZX025D08) (to WZ), and the Emory URC Research Award and the Disruptive Discovery Seed Program by Emory University’s Office of the Provost (to ST). 
The content in this document are those of the authors and should not be interpreted as necessarily representing the views of the funding sources. 
The authors declare no competing interests.
\end{ack}

\bibliographystyle{plainnat}
\bibliography{ref}

\clearpage

\appendix

\crefalias{section}{appendix}
\crefalias{subsection}{appendix}
\vspace{1cm}
\begin{center}
    {\LARGE \textbf{Supplementary Materials}}
\end{center}

\section{Limitations}
\label{sec:limitations}
\textsc{Adwm} has two practical scope conditions. First, like all OPE
estimators, it is bottlenecked by ground-truth quality: on WebShop
the binomial standard error on GT success rate (up to $4.5\%$ with
$128$ episodes) is comparable to the gap between adjacent LEAP iterates
($3$--$19\%$), so finer cross-iterate distinctions require substantially
more GT episodes. Second, the $\psi$ adapter is tied to the evaluation
LLM's embedding space, so each new evaluation LLM requires a short
adapter fit, while neither the offline trajectories nor the world model
need to be recollected or retrained. Neither condition reflects a
fundamental limitation of the underlying formulation.

\section{Relation to Classical Importance Sampling}
\label{app:is_derivation}

We show that the policy-guided target law~\eqref{eq:pgd_target}
is the natural generalization of importance sampling to the setting
where the behavior policy $\pi_b$ is unknown, and that it recovers
the evaluation trajectory law exactly when $p_\theta = p_{\pi_b}$
and $\alpha = 1$.

\paragraph{Classical importance sampling identity.}
For any two policies $\pi_b$ and $\pi_e$, the induced trajectory
laws satisfy:
\begin{align}
p_{\pi_e}(\tau)
&=
\mu(o_1)\prod_{t=1}^{T}\pi_e(a_t\mid h_t)\,P(o_{t+1}\mid h_t,a_t)
\nonumber\\
&=
\mu(o_1)\prod_{t=1}^{T}\pi_b(a_t\mid h_t)\,P(o_{t+1}\mid h_t,a_t)
\cdot
\prod_{t=1}^{T}\frac{\pi_e(a_t\mid h_t)}{\pi_b(a_t\mid h_t)}
\nonumber\\
&=
p_{\pi_b}(\tau)\cdot
\prod_{t=1}^{T}\frac{\pi_e(a_t\mid h_t)}{\pi_b(a_t\mid h_t)},
\label{eq:app_is_identity}
\end{align}
where $\mu(o_1)$ and $P(o_{t+1}\mid h_t,a_t)$ denote the initial
observation law and the observation kernel induced by $\mu$, $P$, and
$\mathcal{E}$ after marginalizing the latent states, and the environment
term $P(o_{t+1}\mid h_t,a_t)$ cancels between numerator and denominator. Taking logarithms,
\begin{align}
\log p_{\pi_e}(\tau)
=
\log p_{\pi_b}(\tau)
+
\sum_{t=1}^{T}\log\pi_e(a_t\mid h_t)
-
\sum_{t=1}^{T}\log\pi_b(a_t\mid h_t).
\label{eq:app_is_log}
\end{align}
Equation~\eqref{eq:app_is_log} is the starting point shared by
importance sampling, policy-guided diffusion~\citep{jackson2024policy},
and STITCH-OPE~\citep{stitch}: the target trajectory law differs from the behavior
trajectory law by a trajectory-level policy-likelihood correction.

\paragraph{Why $\pi_b$ is unavailable.}
In our agentic setting, the offline dataset $\mathcal{D}$ may
aggregate trajectories from heterogeneous sources---different
prompting strategies, decoding configurations, or agent versions.
The behavior policy $\pi_b$ is therefore a mixture whose density
$\pi_b(a_t\mid h_t)$ is generally unavailable in closed form,
preventing direct instantiation of~\eqref{eq:app_is_identity}.

\paragraph{From IS to the policy-guided target law.}
We replace the unknown $p_{\pi_b}(\tau)$ with a learned generative
model $p_\theta(\tau)$ trained on $\mathcal{D}$ by maximum
likelihood. Dropping the negative behavior term
$-\sum_t\log\pi_b(a_t\mid h_t)$ from~\eqref{eq:app_is_log} acts
as a form of behavior regularization~\citep{jackson2024policy}:
it anchors the target distribution to the support of the offline
data rather than allowing unbounded correction away from it.
Exponentiating and introducing a strength parameter $\alpha > 0$
yields:
\begin{align}
q_\alpha(\tau\mid\pi_e)
\;\propto\;
p_\theta(\tau)\,
\exp\!\Big(
\alpha\sum_{t=1}^{T}\log\pi_e(a_t\mid h_t)
\Big),
\label{eq:app_pgd_target}
\end{align}
which is~\eqref{eq:pgd_target} in the main text.

\paragraph{Recovery of $p_{\pi_e}$ at $p_\theta = p_{\pi_b}$, $\alpha=1$.}
Suppose the learned prior is exact, i.e.\ $p_\theta = p_{\pi_b}$.
Then~\eqref{eq:app_pgd_target} with $\alpha = 1$ gives:
\begin{align}
q_1(\tau\mid\pi_e)
&\;\propto\;
p_{\pi_b}(\tau)\,
\exp\!\Big(
\sum_{t=1}^{T}\log\pi_e(a_t\mid h_t)
\Big)
\nonumber\\
&=
p_{\pi_b}(\tau)\,
\prod_{t=1}^{T}\pi_e(a_t\mid h_t).
\label{eq:app_recovery_unnorm}
\end{align}
The normalizing constant of~\eqref{eq:app_recovery_unnorm} is:
\begin{align}
Z
&=
\int p_{\pi_b}(\tau)\,\prod_{t=1}^{T}\pi_e(a_t\mid h_t)\,d\tau
\nonumber\\
&=
\int
\mu(o_1)\prod_{t=1}^{T}\pi_b(a_t\mid h_t)\,P(o_{t+1}\mid h_t,a_t)
\cdot
\prod_{t=1}^{T}\pi_e(a_t\mid h_t)
\,d\tau.
\label{eq:app_Z_start}
\end{align}
Using~\eqref{eq:app_is_identity} to substitute
$p_{\pi_b}(\tau)\prod_t\pi_e(a_t\mid h_t)/\pi_b(a_t\mid h_t)
= p_{\pi_e}(\tau)$:
\begin{align}
Z
&=
\int
p_{\pi_b}(\tau)\,
\prod_{t=1}^{T}
\frac{\pi_e(a_t\mid h_t)}{\pi_b(a_t\mid h_t)}
\cdot
\prod_{t=1}^{T}\pi_b(a_t\mid h_t)
\,d\tau
\nonumber\\
&=
\int p_{\pi_e}(\tau)\,d\tau
\;=\;
1.
\label{eq:app_Z_equals_one}
\end{align}
Therefore~\eqref{eq:app_recovery_unnorm} is already normalized, and:
\begin{align}
q_1(\tau\mid\pi_e)
\Big|_{p_\theta = p_{\pi_b}}
\;=\;
p_{\pi_b}(\tau)\,\prod_{t=1}^{T}\pi_e(a_t\mid h_t)
\;=\;
p_{\pi_e}(\tau),
\label{eq:app_exact_recovery}
\end{align}
where the last equality follows from~\eqref{eq:app_is_identity}.
This confirms that $q_\alpha$ with $\alpha=1$ and $p_\theta =
p_{\pi_b}$ recovers the evaluation trajectory law exactly.

\paragraph{Role of $\alpha$.}
For $\alpha \neq 1$ or $p_\theta \neq p_{\pi_b}$,
$q_\alpha(\tau\mid\pi_e)$ interpolates between the offline prior
($\alpha\to 0$, recovering $p_\theta$) and a distribution that
increasingly concentrates on trajectories preferred by $\pi_e$
($\alpha\to\infty$). In practice, $\alpha$ is a hyperparameter
that balances the fidelity of the simulated environment against
the alignment of generated rollouts with the evaluation policy.
\section{Proof of Proposition~\ref{prop:ar_projection}}
\label{app:proof}

We derive the single-step conditional law for $o_{t+1}$
induced by the policy-guided trajectory law
$q_\alpha(\tau \mid \pi_e)$ defined in
\eqref{eq:pgd_target}.

Fix a history $h_t = (o_1, a_1, \dots, o_t)$ and a
current action $a_t$. We wish to characterize the
conditional distribution $q_\alpha(o_{t+1} \mid h_t,
a_t; \pi_e)$. Denote the future suffix trajectory as:
\[
\tau_{t+1:T} = (o_{t+1}, a_{t+1}, o_{t+2},
\dots, a_T, o_{T+1}),
\]
so that the full trajectory decomposes as
$\tau = (h_t, a_t, \tau_{t+1:T})$.

By definition of $q_\alpha$ in \eqref{eq:pgd_target},
the unnormalized density of any trajectory $\tau$ is:
\[
q_\alpha(\tau \mid \pi_e)
\propto
p_\theta(\tau)\,
\exp\!\Bigl(
\alpha \sum_{t'=1}^{T} \log \pi_e(a_{t'} \mid h_{t'})
\Bigr).
\]
Conditioning on $(h_t, a_t)$ amounts to fixing all
variables up to and including $a_t$. The terms
$\sum_{t'=1}^{t} \log \pi_e(a_{t'} \mid h_{t'})$ in the
exponent are therefore constant with respect to the
future suffix $\tau_{t+1:T}$, and are absorbed into the
normalizing constant. We obtain:
\begin{align}
q_\alpha(\tau_{t+1:T} \mid h_t, a_t; \pi_e)
&= \frac{
q_\alpha(\tau \mid \pi_e)
}{
q_\alpha(h_t, a_t \mid \pi_e)
}
\nonumber \\
&\propto
p_\theta(\tau_{t+1:T} \mid h_t, a_t)\,
\exp\!\Bigl(
\alpha \sum_{u=t+1}^{T} \log \pi_e(a_u \mid h_u)
\Bigr),
\label{eq:suffix_cond_proof}
\end{align}
where we have used the fact that
$p_\theta(\tau) = p_\theta(h_t, a_t) \cdot
p_\theta(\tau_{t+1:T} \mid h_t, a_t)$
by the chain rule of probability.

We now factor $p_\theta(\tau_{t+1:T} \mid h_t, a_t)$
by separating out $o_{t+1}$ from the remaining suffix.
By the chain rule:
\begin{align}
p_\theta(\tau_{t+1:T} \mid h_t, a_t)
= p_\theta(o_{t+1} \mid h_t, a_t)\,
p_\theta(\tau_{t+2:T} \mid h_t, a_t, o_{t+1}),
\label{eq:chain_rule_proof}
\end{align}
where we write $\tau_{t+2:T} = (a_{t+1}, o_{t+2}, \dots,
a_T, o_{T+1})$ for the suffix starting after $o_{t+1}$.
Note that once $o_{t+1}$ is fixed, the updated history
is $h_{t+1} = (h_t, a_t, o_{t+1})$, so
$p_\theta(\tau_{t+2:T} \mid h_t, a_t, o_{t+1})
= p_\theta(\tau_{t+2:T} \mid h_{t+1})$.

Substituting \eqref{eq:chain_rule_proof} into
\eqref{eq:suffix_cond_proof}:
\begin{align}
q_\alpha(\tau_{t+1:T} \mid h_t, a_t; \pi_e)
\propto
p_\theta(o_{t+1} \mid h_t, a_t)\,
p_\theta(\tau_{t+2:T} \mid h_{t+1})\,
\exp\!\Bigl(
\alpha \sum_{u=t+1}^{T} \log \pi_e(a_u \mid h_u)
\Bigr).
\label{eq:factored_proof}
\end{align}

To obtain the marginal over $o_{t+1}$, we integrate
\eqref{eq:factored_proof} over all remaining future
variables $\tau_{t+2:T}$:
\begin{align}
q_\alpha(o_{t+1} \mid h_t, a_t; \pi_e)
&= \int
q_\alpha(\tau_{t+1:T} \mid h_t, a_t; \pi_e)
\, d\tau_{t+2:T}
\nonumber \\
&\propto
p_\theta(o_{t+1} \mid h_t, a_t)
\int
p_\theta(\tau_{t+2:T} \mid h_{t+1})\,
\exp\!\Bigl(
\alpha \sum_{u=t+1}^{T} \log \pi_e(a_u \mid h_u)
\Bigr)
\, d\tau_{t+2:T}.
\label{eq:marginal_proof}
\end{align}
The integral in \eqref{eq:marginal_proof} can be
recognized as an expectation under the prior over
future trajectories rooted at $o_{t+1}$. Specifically,
since $p_\theta(\tau_{t+2:T} \mid h_{t+1})$ is a
probability distribution over $\tau_{t+2:T}$:
\begin{align}
&\int
p_\theta(\tau_{t+2:T} \mid h_{t+1})\,
\exp\!\Bigl(
\alpha \sum_{u=t+1}^{T} \log \pi_e(a_u \mid h_u)
\Bigr)
\, d\tau_{t+2:T}
\nonumber \\
&=
\int
p_\theta(\tau_{t+1:T} \mid h_t, a_t, o_{t+1})\,
\exp\!\Bigl(
\alpha \sum_{u=t+1}^{T} \log \pi_e(a_u \mid h_u)
\Bigr)
\, d\tau_{t+1:T}
\nonumber \\
&=
\mathbb{E}_{\tau_{t+1:T} \sim
p_\theta(\cdot \mid h_t, a_t, o_{t+1})}
\!\left[
\exp\!\Bigl(
\alpha \sum_{u=t+1}^{T}
\log \pi_e(a_u \mid h_u)
\Bigr)
\right],
\label{eq:expectation_proof}
\end{align}
where in the second equality we have re-expressed the
integral as an expectation over the full future suffix
$\tau_{t+1:T}$ conditioned on $(h_t, a_t, o_{t+1})$,
noting that integrating over $\tau_{t+2:T}$ under
$p_\theta(\cdot \mid h_{t+1})$ is equivalent to
integrating over $\tau_{t+1:T}$ under
$p_\theta(\cdot \mid h_t, a_t, o_{t+1})$ since the
$o_{t+1}$ component is fixed in both cases.

The expectation in \eqref{eq:expectation_proof} is
precisely the continuation factor
$\mathcal{C}_\alpha(o_{t+1}; h_t, a_t, \pi_e)$ defined
in \eqref{eq:continuation}. Substituting back into
\eqref{eq:marginal_proof}:
\begin{align}
q_\alpha(o_{t+1} \mid h_t, a_t; \pi_e)
\propto
p_\theta(o_{t+1} \mid h_t, a_t)\,
\mathcal{C}_\alpha(o_{t+1}; h_t, a_t, \pi_e).
\end{align}
This is precisely \eqref{eq:ar_projection},
completing the proof. \hfill$\square$

\section{Proof of Theorem~\ref{thm:fused}}
\label{app:fusion}

We derive the fused single-step conditional law
\eqref{eq:fused} from Proposition~\ref{prop:ar_projection}.

By Proposition~\ref{prop:ar_projection}, the single-step
conditional induced by $q_\alpha$ satisfies:
\begin{equation}
q_{\alpha}(o_{t+1}\mid h_{t},a_{t};\pi_{e})
\propto
p_{\theta}(o_{t+1}\mid h_{t},a_{t})\,
\mathcal{C}_{\alpha}(o_{t+1};h_{t},a_{t},\pi_{e}).
\label{eq:ar_proj_app}
\end{equation}
It remains to decompose $p_\theta(o_{t+1} \mid h_t, a_t)$.
Consider the joint distribution $p_\theta(o_{t+1}, a_t
\mid h_t)$. By the chain rule, this factors in two ways:
\begin{align}
p_\theta(o_{t+1}, a_t \mid h_t)
&= p_\theta(o_{t+1} \mid h_t, a_t)\,
p_\theta(a_t \mid h_t),
\label{eq:joint_a} \\
p_\theta(o_{t+1}, a_t \mid h_t)
&= p_\theta(a_t \mid o_{t+1}, h_t)\,
p_\theta(o_{t+1} \mid h_t).
\label{eq:joint_b}
\end{align}
Equating \eqref{eq:joint_a} and \eqref{eq:joint_b}:
\begin{equation}
p_\theta(o_{t+1} \mid h_t, a_t)\,
p_\theta(a_t \mid h_t)
= p_\theta(a_t \mid o_{t+1}, h_t)\,
p_\theta(o_{t+1} \mid h_t).
\end{equation}
Dividing both sides by $p_\theta(a_t \mid h_t) > 0$:
\begin{equation}
p_\theta(o_{t+1} \mid h_t, a_t)
= \frac{p_\theta(a_t \mid o_{t+1}, h_t)\,
p_\theta(o_{t+1} \mid h_t)}{p_\theta(a_t \mid h_t)}.
\label{eq:bayes_app}
\end{equation}
Since $p_\theta(a_t \mid h_t)$ is constant with respect
to $o_{t+1}$ for fixed $(h_t, a_t)$, \eqref{eq:bayes_app}
gives:
\begin{equation}
p_\theta(o_{t+1} \mid h_t, a_t)
\propto
p_\theta(o_{t+1} \mid h_t)\,
p_\theta(a_t \mid o_{t+1}, h_t).
\label{eq:bayes_propto}
\end{equation}
Substituting \eqref{eq:bayes_propto} into
\eqref{eq:ar_proj_app}:
\begin{equation}
q_\alpha(o_{t+1} \mid h_t, a_t; \pi_e)
\propto
p_\theta(o_{t+1} \mid h_t)\,
p_\theta(a_t \mid o_{t+1}, h_t)\,
\mathcal{C}_\alpha(o_{t+1}; h_t, a_t, \pi_e),
\end{equation}
which is the $\omega = \eta = 1$ case of
\eqref{eq:fused}.

For general $\omega > 0$ and $\eta \geq 0$, we introduce
the steered family by replacing each factor with its
$\omega$-th and $\eta$-th power respectively. This
corresponds to treating $\omega$ and $\eta$ as inverse
temperatures on the action posterior and continuation
factor, giving:
\begin{equation}
\widetilde{q}_{\omega,\eta}(o_{t+1} \mid h_t, a_t; \pi_e)
\propto
p_\theta(o_{t+1} \mid h_t)\,
p_\theta(a_t \mid o_{t+1}, h_t)^\omega\,
\mathcal{C}_\alpha(o_{t+1}; h_t, a_t, \pi_e)^\eta.
\end{equation}
This is \eqref{eq:fused}, completing the proof.
\hfill$\square$

\begin{remark}
The denominator $p_\theta(a_t \mid h_t)$ in
\eqref{eq:bayes_app} depends only on $(h_t, a_t)$ and
not on $o_{t+1}$. It is therefore a constant with respect
to the variable of integration and is absorbed into the
normalizing constant of the conditional distribution.
\end{remark}

\begin{remark}
The steered family $\widetilde{q}_{\omega,\eta}$ is a
valid probability distribution for any $\omega > 0$ and
$\eta \geq 0$, provided the normalizing constant
\[
Z(\omega, \eta; h_t, a_t) = \int
p_\theta(o_{t+1} \mid h_t)\,
p_\theta(a_t \mid o_{t+1}, h_t)^\omega\,
\mathcal{C}_\alpha(o_{t+1}; h_t, a_t, \pi_e)^\eta
\, do_{t+1}
\]
is finite, which holds under standard regularity
conditions on $p_\theta$ and $\pi_e$.
\end{remark}
\section{Derivation of the Guided Score Function}
\label{app:score_derivation}

We derive \eqref{eq:final_score} in three steps, using the notation of
\cref{sec:guided_diffusion}: (i) decompose the
score of the fused conditional into three additive terms; (ii) show
that the first two terms are realized by classifier-free guidance;
(iii) derive the tractable approximation of the continuation score.

\subsection{Score decomposition of the fused conditional}

By Theorem~\ref{thm:fused}, the target single-step conditional is:
\begin{align}
\widetilde{q}_{\omega,\eta}(z_{t+1}\mid h_t^z,a_t;\pi_e)
\;\propto\;
p_{\theta}(z_{t+1}\mid h_t^z)^{1}
\cdot
p_{\phi}(a_t\mid z_{t+1},h_t^z)^{\omega}
\cdot
\mathcal{C}_{\alpha}(z_{t+1};h_t^z,a_t,\pi_e)^{\eta}.
\label{eq:app_fused}
\end{align}
Taking logarithms:
\begin{align}
\log\widetilde{q}_{\omega,\eta}(z_{t+1}\mid h_t^z,a_t;\pi_e)
=
\log p_{\theta}(z_{t+1}\mid h_t^z)
+\omega\log p_{\phi}(a_t\mid z_{t+1},h_t^z)
+\eta\log\mathcal{C}_{\alpha}(z_{t+1};h_t^z,a_t,\pi_e)
+\text{const},
\label{eq:app_log_fused}
\end{align}
where the constant absorbs the normalizer, which does not depend on
$z_{t+1}$. Differentiating with respect to $z_{t+1}$ gives the exact
score:
\begin{align}
\nabla_{z_{t+1}}\log\widetilde{q}_{\omega,\eta}
=
\underbrace{\nabla_{z_{t+1}}\log p_{\theta}(z_{t+1}\mid h_t^z)}_{\text{prior score}}
+\omega\underbrace{\nabla_{z_{t+1}}\log p_{\phi}(a_t\mid z_{t+1},h_t^z)}_{\text{action-posterior score}}
+\eta\underbrace{\nabla_{z_{t+1}}\log\mathcal{C}_{\alpha}(z_{t+1};h_t^z,a_t,\pi_e)}_{\text{continuation score}}.
\label{eq:app_score_decomp}
\end{align}
Each term has a distinct role: the prior score keeps generated latents
on the manifold of the offline data; the action-posterior score steers
generation toward latents from which $a_t$ is recoverable; the
continuation score steers generation toward latents that open up
futures compatible with $\pi_e$.

\subsection{Realization of the prior and action-posterior scores via
classifier-free guidance}

We show that the first two terms of \eqref{eq:app_score_decomp} are
jointly realized by classifier-free guidance~\citep{ho2022classifier}
without a separate classifier.

Consider the joint density $p_\theta(z_{t+1}, a_t \mid h_t^z)$.
By the chain rule it factors in two ways:
\begin{align}
p_\theta(z_{t+1}, a_t \mid h_t^z)
&= p_\theta(z_{t+1} \mid h_t^z, a_t)\,p_\theta(a_t \mid h_t^z),
\label{eq:app_chain_a}\\
p_\theta(z_{t+1}, a_t \mid h_t^z)
&= p_\phi(a_t \mid z_{t+1}, h_t^z)\,p_\theta(z_{t+1} \mid h_t^z).
\label{eq:app_chain_b}
\end{align}
Equating \eqref{eq:app_chain_a} and \eqref{eq:app_chain_b} and
dividing both sides by $p_\theta(a_t \mid h_t^z) > 0$:
\begin{align}
p_\theta(z_{t+1} \mid h_t^z, a_t)
= \frac{p_\phi(a_t \mid z_{t+1}, h_t^z)\,
        p_\theta(z_{t+1} \mid h_t^z)}
       {p_\theta(a_t \mid h_t^z)}.
\label{eq:app_bayes}
\end{align}
Taking logarithms and differentiating with respect to $z_{t+1}$,
noting that $p_\theta(a_t \mid h_t^z)$ is constant in $z_{t+1}$:
\begin{align}
\nabla_{z_{t+1}}\log p_\theta(z_{t+1} \mid h_t^z, a_t)
=
\nabla_{z_{t+1}}\log p_\theta(z_{t+1} \mid h_t^z)
+
\nabla_{z_{t+1}}\log p_\phi(a_t \mid z_{t+1}, h_t^z).
\label{eq:app_score_bayes}
\end{align}
Rearranging:
\begin{align}
\nabla_{z_{t+1}}\log p_\phi(a_t \mid z_{t+1}, h_t^z)
=
\nabla_{z_{t+1}}\log p_\theta(z_{t+1} \mid h_t^z, a_t)
-
\nabla_{z_{t+1}}\log p_\theta(z_{t+1} \mid h_t^z).
\label{eq:app_posterior_score}
\end{align}
Substituting \eqref{eq:app_posterior_score} into the first two terms
of \eqref{eq:app_score_decomp}:
\begin{align}
&\nabla_{z_{t+1}}\log p_\theta(z_{t+1}\mid h_t^z)
+\omega\,\nabla_{z_{t+1}}\log p_\phi(a_t\mid z_{t+1},h_t^z)
\nonumber\\
&=
\nabla_{z_{t+1}}\log p_\theta(z_{t+1}\mid h_t^z)
+\omega\Big(
\nabla_{z_{t+1}}\log p_\theta(z_{t+1}\mid h_t^z,a_t)
-\nabla_{z_{t+1}}\log p_\theta(z_{t+1}\mid h_t^z)
\Big)
\nonumber\\
&=
(1-\omega)\,\nabla_{z_{t+1}}\log p_\theta(z_{t+1}\mid h_t^z)
+\omega\,\nabla_{z_{t+1}}\log p_\theta(z_{t+1}\mid h_t^z,a_t).
\label{eq:app_cfg_derivation}
\end{align}
Writing $\omega = 1+\lambda$, \eqref{eq:app_cfg_derivation} becomes:
\begin{align}
(1+\lambda)\,\nabla_{z_{t+1}}\log p_\theta(z_{t+1}\mid h_t^z,a_t)
-\lambda\,\nabla_{z_{t+1}}\log p_\theta(z_{t+1}\mid h_t^z),
\label{eq:app_cfg_final}
\end{align}
which is exactly the classifier-free guidance combination with scale
$\lambda$. In practice, both scores are approximated by the
action-conditioned and unconditional branches of the shared score
network $s_\theta$, trained jointly via \eqref{eq:dsm_loss} with
action dropout.

\subsection{Tractable approximation of the continuation score}

The continuation factor is defined as:
\begin{align}
\mathcal{C}_\alpha(z_{t+1};h_t^z,a_t,\pi_e)
:=
\mathbb{E}_{\tau_{t+1:T}\sim p_\theta(\cdot\mid h_t^z,a_t,z_{t+1})}
\!\left[
\exp\!\Big(
\alpha\sum_{u=t+1}^{T}\log\pi_e(a_u\mid h_u^\pi)
\Big)
\right].
\label{eq:app_continuation}
\end{align}
Computing $\nabla_{z_{t+1}}\log\mathcal{C}_\alpha$ exactly requires
differentiating through an expectation over all future trajectories,
which is intractable. We derive a tractable approximation via the
denoised estimate.

At diffusion step $k$, the score network provides the posterior mean
estimate:
\begin{align}
\hat{z}_{t+1}
= \mathbb{E}[z_{t+1} \mid z_{t+1}^k]
= \frac{z_{t+1}^k + (1-\bar{\alpha}_k)\,
        s_\theta(z_{t+1}^k, k; h_t^z, a_t)}
       {\sqrt{\bar{\alpha}_k}}.
\label{eq:app_denoised}
\end{align}
We substitute $\hat{z}_{t+1}$ for $z_{t+1}$ in
\eqref{eq:app_continuation} and replace the expectation over future
trajectories with a single greedy rollout
$\hat{\tau}_{t+1:T} = \mathcal{R}_\theta(\hat{z}_{t+1};h_t^z,a_t)$:
\begin{align}
\log\mathcal{C}_\alpha(z_{t+1};h_t^z,a_t,\pi_e)
\;\approx\;
\alpha\sum_{u=t+1}^{T}
\log\pi_e(\hat{a}_u\mid h_u^\pi).
\label{eq:app_cont_approx}
\end{align}
Differentiating \eqref{eq:app_cont_approx} with respect to
$\hat{z}_{t+1}$, and using this gradient in place of the gradient with
respect to $z_{t+1}^k$ (that is, dropping the Jacobian of
\eqref{eq:app_denoised}), gives the tractable continuation gradient:
\begin{align}
\eta\,\nabla_{z_{t+1}}\log\mathcal{C}_\alpha
\;\approx\;
\eta_k\,\alpha\,
\nabla_{\hat{z}_{t+1}}
\sum_{u=t+1}^{T}
\log\pi_e(\hat{a}_u\mid h_u^\pi),
\label{eq:app_cont_grad}
\end{align}
where we write $\eta_k$ to allow the continuation weight to vary
with noise level $k$, suppressing the signal when $\hat{z}_{t+1}$
is unreliable at high noise and amplifying it late in the reverse
process.

\subsection{Final guided score}

Combining \eqref{eq:app_cfg_final} and \eqref{eq:app_cont_grad},
the complete guided score at noise level $k$ is:
\begin{align}
s(z_{t+1}^{k},k;h_t^z,a_t,\pi_e)
&=
\nabla_{z_{t+1}^{k}}
\log p_{\theta}(z_{t+1}^{k}\mid h_t^z)
+
\omega\,
\nabla_{z_{t+1}^{k}}
\log p_{\phi}(a_t\mid z_{t+1}^{k},h_t^z)
\nonumber\\
&\quad
+\;\eta_k\,\alpha\,
\nabla_{\hat{z}_{t+1}}
\sum_{u=t+1}^{T}
\log \pi_e(\hat{a}_u\mid h_u^\pi),
\label{eq:app_final_score}
\end{align}
which is \eqref{eq:final_score} in the main text.

\section{Experimental Setup Details}
\label{app:setup}

\subsection{Software stack and reproducibility}
All training and OPE rollout code is implemented in PyTorch~2.6 with
Transformers~4.56 under Python~3.10. LLM-side $\pi_b$/$\pi_e$ inference
is served by vLLM~0.8.5 with bf16 weights and a tensor-parallel size of
1 (one GPU per LLM job). Baselines (FQE, DR, IS, WIS, DM) are
implemented to match the COBS reference protocol of
\citet{voloshin2019empirical} verbatim except for the per-$\varepsilon$
FQE correction described in \cref{app:fqe-correction}. Random seeds
$s\in\{0,1,2,3,4\}$ control
(i) the $\varepsilon$-greedy admissible-action draw and
(ii) the imagined-rollout DDPM noise sequence. The world model and the
$\psi$ adapter are trained \emph{once per behavior pool} with a fixed
seed; the five per-cell seeds vary inference-time stochasticity only.

\subsection{Hardware and compute budget}
All experiments run on a shared cluster node with $2\times$\,Intel
Xeon Gold 6248R (3.00\,GHz, 96 logical cores) and $2\times$\,NVIDIA
Quadro RTX~8000 (48\,GB) GPUs interconnected by PCIe~3.0~x16. We
allocate a single GPU per training or evaluation job; multi-GPU
parallelism is not used.

\paragraph{Wall-clock budget.}
\begin{itemize}
  \item World-model training (per environment): 50 epochs on
        256--744 trajectories at batch size 64 takes 30--90\,min.
  \item $\psi$-adapter training (per environment): 20 epochs at
        batch size 8 takes 10--20\,min.
  \item GT collection (per cell, per $\varepsilon$): vLLM-served
        $\pi_e^\varepsilon$ rolling out for 64--128 episodes takes
        10--40\,min depending on env horizon.
  \item \textsc{Adwm} inference (per cell, per seed): 5 $\varepsilon$ levels
        $\times$ 32 imagined rollouts $\times$ horizon $\le 30$ takes
        15--45\,min.
  \item FQE / DR (per cell, per seed): 5 $\varepsilon$-conditioned
        Q-nets at 20 epochs each takes 5--10\,min.
\end{itemize}

\paragraph{Total cost.}
The full camera-ready experimental suite (six $(\pi_b,\pi_e)$ cells
$\times$ 5 seeds, plus 7 ablation variants on three environments,
plus 5 baseline estimators per cell) consumes approximately
$300$ GPU-hours, of which $\sim$$70\%$ is LLM-side ($\pi_b$ and $\pi_e$)
inference and $\sim$$25\%$ is GT collection. World-model training
itself accounts for $<5\%$ of the budget --- the dominant cost of
\textsc{Adwm} is the same as the dominant cost of any LLM-agent OPE method,
namely running the eval LLM.

\subsection{Behavior-policy data}
For each $(\pi_b,\pi_e)$ cell we collect a single offline behavior
pool that is reused across all $\varepsilon$ levels and seeds for
that cell. The world model is trained \emph{only} on this pool.
\Cref{tab:behdata} summarises the per-cell trajectory counts,
behavior models actually used, and observed empirical statistics.
The HotpotQA, ScienceWorld, and ALFWorld-iter1 cells use a single
behavior LLM each. The ALFWorld-iter3 and WebShop-iter1 cells use a
three-LLM behavior pool consisting of the generic
Llama-3.1-8B-Instruct together with two LEAP iterates ($\text{iter}_0$
and $\text{iter}_2$); we exclude $\text{iter}_1$ from the WS pool and
$\text{iter}_3$ from the ALF pool to ensure the eval policy is never
in the behavior pool. This pooling choice mirrors a realistic
deployment in which multiple released checkpoints from the same LEAP
self-improvement chain may already be available offline. We verify
post-hoc that no training trajectory in the pool was generated by the
exact $\pi_e$ checkpoint scored at evaluation time.

\begin{table}[h]
\centering
\small
\setlength{\tabcolsep}{4pt}
\renewcommand{\arraystretch}{1.15}
\caption{Behavior pool composition and statistics, per cell. ``Avg.\
horizon'' is the mean number of environment steps per $\pi_b$
trajectory; ``$\pi_b$ succ.\ rate'' is the empirical success rate on
the collected pool.}
\label{tab:behdata}
\begin{tabular}{@{}lllcc@{}}
\toprule
Cell & $\pi_b$ pool & $N$ traj.\ & Avg.\ horizon & $\pi_b$ succ.\ rate \\
\midrule
HotpotQA-DPO/PRM/cross & ReAct-HotpotQA-SFT (single)            & 512 &  6.4 & 12.5\% \\
ScienceWorld-ETO       & sw-llama-sft (single)                  & 512 &  9.1 & 11.7\% \\
ALFWorld-iter1         & Llama-3.1-8B-Instruct (single)         & 256 & 29.5 &  3.5\% \\
ALFWorld-iter3         & Llama-3.1-8B-Instr.\ + leap-alf-iter\{0,2\} & 744 & 29.2 &  5.8\% \\
WebShop-iter1          & Llama-3.1-8B-Instr.\ + leap-ws-iter\{0,2\}  & 640 & 14.4 &  5.0\% \\
\bottomrule
\end{tabular}
\end{table}

Trajectories are collected with greedy LLM decoding (temperature 0)
and two system-prompt variants $\{p_0,p_1\}$ that differ only in
instruction phrasing (not task content), to obtain useful behavior
coverage without changing the underlying $\pi_b$. Episode length is
capped at 30 environment steps for ALFWorld and WebShop and at 12 for
HotpotQA and ScienceWorld. Each recorded trajectory contains, per step,
\[
(\text{obs\_text}_t,\text{llm\_response}_t,u_t,a_t,
a_{\text{valid}}^t,r_t,\text{done}_t),
\]
where \(u_t\) is the raw action text parsed from the LLM response and
\(a_t\) is the action actually passed to the environment after the
environment-specific projection interface. We write this interface as
\[
a_t\sim
\Gamma_{\mathrm{env}}
\!\left(
\cdot\mid h_t,u_t,\mathcal A_{\mathrm{adm}}(h_t)
\right).
\]
The world-model transition tuples use the executed action \(a_t\), so
\((o_{t+1},r_t)\) is paired with the action that generated it. The raw
response and \(u_t\) are retained as auxiliary logged text. Successful
trajectories are upsampled \(5\times\) during world-model training to
counter class imbalance on the long-tail success distributions of
\cref{tab:behdata}.

\subsection{Ground-truth episode counts}
Per-cell ground-truth success curves are obtained by running
$\pi_e^\varepsilon$ in the actual environment for $n_{\text{GT}}$
independent episodes per $\varepsilon$ level, with seed 42 fixed
across $\varepsilon$ within a policy:

\begin{center}
\small
\begin{tabular}{@{}lc@{}}
\toprule
Cell & $n_{\text{GT}}$ per $\varepsilon$ \\
\midrule
HotpotQA-DPO   & 128 \\
HotpotQA-PRM   & \phantom{0}64 \\
ScienceWorld-ETO & \phantom{0}64 \\
ALFWorld-iter1 / iter3 & \phantom{0}72 \\
WebShop-iter1  & 128 \\
\bottomrule
\end{tabular}
\end{center}

The binomial standard error on the GT success rate at $n{=}64$ is
$\le 6.3\%$ for any $p\in(0,1)$, and $\le 4.5\%$ at $n{=}128$.

\subsection{$\varepsilon$-greedy mixing protocol}
Let \(\widetilde\pi_e(u\mid h)\) denote the evaluation LLM's distribution
over raw action text and let
\(\Gamma_{\mathrm{env}}(a\mid h,u,\mathcal A_{\mathrm{adm}}(h))\)
denote the environment-specific projection from raw text to an executed
admissible action. The induced executed-action policy is
\begin{equation}
\pi_{e,\Gamma}(a\mid h)
=
\sum_u
\widetilde\pi_e(u\mid h)\,
\Gamma_{\mathrm{env}}
\!\left(
a\mid h,u,\mathcal A_{\mathrm{adm}}(h)
\right).
\label{eq:projected-policy}
\end{equation}
This notation also covers a deterministic projection, in which
\(\Gamma_{\mathrm{env}}\) is a point mass.

We apply \(\varepsilon\)-mixing to the executed-action policy:
\begin{equation}
\pi_e^\varepsilon(a\mid h)
=
(1-\varepsilon)\pi_{e,\Gamma}(a\mid h)
+
\varepsilon
\frac{
\mathbb I[a\in\mathcal A_{\mathrm{adm}}(h)]
}{
|\mathcal A_{\mathrm{adm}}(h)|
}.
\label{eq:eps-mix}
\end{equation}
Operationally, we sample one Bernoulli variable per step. With probability
\(1-\varepsilon\), the LLM produces \(u_t\sim\widetilde\pi_e(\cdot\mid h_t)\)
and the interface samples
\(a_t\sim\Gamma_{\mathrm{env}}(\cdot\mid h_t,u_t,
\mathcal A_{\mathrm{adm}}(h_t))\). With probability \(\varepsilon\), we
sample \(a_t\) uniformly from the current admissible set. Both
\textsc{Adwm} imagined rollouts and ground-truth environment rollouts use
this same interface and random seed.

\subsection{World-model architecture}
\label{app:wm-arch}
The world model is a 7.4M-parameter latent diffusion model with the
modular structure shown in \cref{tab:wm-arch}. In this appendix we write
$f_\theta$, $\bar h_t$, $\epsilon_\theta$, $R_\theta$, and $D_\theta$ for
the encoder $E$, the history summary of $h_t^z$, the noise-prediction form
of the score network $s_\theta$, the reward head $r_\rho$, and the
termination head $d_\rho$ of \cref{sec:guided_diffusion}. The largest component
is the text-based observation encoder (5.05\,M params), which
tokenises raw $\text{obs\_text}_t$ via a stable hash-bucket
tokeniser (vocabulary size 32{,}768, max length 128 tokens) and
embeds it into the latent dimension $d_z{=}64$ through a 4-layer
TransformerEncoder of width 128 with 4 attention heads. The history
encoder is a causal 4-layer Transformer with width 128 and a
generated subsequent-token mask, conditioning on
$(z_{t}, a_{t})$ tuples. The score network is a 3-layer FiLM-MLP of
width 256 with sinusoidal time embedding (dimension 64) and
zero-initialised output projection; the IDM and BC heads are 2-layer
MLPs over $(z, h)$. All non-linearities are SiLU except the IDM /
policy / projector heads, which use ReLU and GELU respectively.

\begin{table}[h]
\centering
\small
\setlength{\tabcolsep}{6pt}
\renewcommand{\arraystretch}{1.15}
\caption{Per-module parameter counts in the \textsc{Adwm} world model
(7.38M parameters total for $d_z{=}64$).}
\label{tab:wm-arch}
\begin{tabular}{@{}llrr@{}}
\toprule
Module & Function & Params (M) & \% of total \\
\midrule
Observation encoder ($f_\theta$) & Text $\to z\in\mathbb{R}^{d_z}$  & 5.053 & 68.4\% \\
History encoder ($\bar h$)       & Causal Transformer over $(z,a)$ pairs   & 0.814 & 11.0\% \\
Soft-token projector             & $z\to\mathbb{R}^{n_{\text{soft}}\times d_{\text{llm}}}$ for $\psi$ adapter & 0.609 & \phantom{0}8.2\% \\
Score net ($\epsilon_\theta$)    & FiLM-MLP for diffusion denoising  & 0.527 & \phantom{0}7.1\% \\
BC policy head                   & $z,h\to a$ logits                 & 0.170 & \phantom{0}2.3\% \\
Reward head ($R_\theta$)         & 2-layer MLP $\to\sigma(\cdot)$    & 0.115 & \phantom{0}1.6\% \\
Termination head ($D_\theta$)    & 2-layer MLP $\to$ BCE logit       & 0.050 & \phantom{0}0.7\% \\
Inverse-dynamics head (IDM)      & $z_t,z_{t+1},h\to a_t$            & 0.044 & \phantom{0}0.6\% \\
Action embedding                 & $\mathbb{Z}\to\mathbb{R}^{d_a}$, $|\mathcal{A}|=256,\,d_a{=}32$ & 0.001 & $<\!0.1\%$ \\
\bottomrule
\end{tabular}
\end{table}

The action vocabulary \( |\mathcal{A}| \) is set per environment from
the unique projected actions \(a_t\) actually executed in the behavior
pool, capped at 256; HotpotQA uses a fixed 4-action vocabulary
$\{\langle\text{unk}\rangle,\,\text{search},\,\text{lookup},\,\text{finish}\}$
that matches the ReAct protocol of \citet{yao2022react}.

\subsection{World-model training objective}
\label{app:wm-loss}
The world model is trained with the multi-task objective
\begin{equation}
\mathcal{L}_{\text{WM}} \;=\;
\mathcal{L}_{\text{DSM}}
+ \lambda_{\text{idm}}\,\mathcal{L}_{\text{IDM}}
+ \lambda_{\text{bc}}\,\mathcal{L}_{\text{BC}}
+ \lambda_{\text{div}}\,\mathcal{L}_{\text{div}}
+ \lambda_{\text{r}}\,\mathcal{L}_{\text{r}}
+ \lambda_{\text{d}}\,\mathcal{L}_{\text{d}},
\label{eq:wm-loss}
\end{equation}
with weights
$(\lambda_{\text{idm}},\lambda_{\text{bc}},\lambda_{\text{div}},\lambda_{\text{r}},\lambda_{\text{d}})
=(0.1,\,0.1,\,0.05,\,10,\,1)$. The component losses are defined as
follows.

\paragraph{(a) Denoising score matching ($\mathcal{L}_{\text{DSM}}$).}
For a sampled diffusion step $k\!\sim\!\mathrm{Unif}(1,K)$, latent
$z\!\sim\!f_\theta(\text{obs}_{t+1})$, history $\bar h_t$, and action
$a_t$, we form the noised latent
$z_k = \sqrt{\bar\alpha_k}\,z + \sqrt{1-\bar\alpha_k}\,\xi$ with
$\xi\!\sim\!\mathcal{N}(0,I)$ and a cosine $\bar\alpha_k$ schedule.
We train the conditional score network $\epsilon_\theta(z_k,k,\bar h_t,a_t)$
together with its unconditional counterpart $\epsilon_\theta(z_k,k,\bar h_t,\varnothing)$
under classifier-free dropout $p_{\text{uncond}}{=}0.1$:
\begin{equation}
\mathcal{L}_{\text{DSM}}\;=\;
\mathbb{E}_{k,\xi}\big[w(k)\,\big(\|\epsilon_\theta(z_k,k,\bar h_t,a_t) - \xi\|_2^2
+ \|\epsilon_\theta(z_k,k,\bar h_t,\varnothing) - \xi\|_2^2\big)\big],
\quad w(k)\,=\,1-\bar\alpha_k.
\end{equation}
The $w(k)\!=\!1-\bar\alpha_k$ weighting keeps the per-step loss
numerically comparable across $k$. This is the noise-prediction form of
\eqref{eq:dsm_loss}, with
$s_\theta=-\epsilon_\theta/\sqrt{1-\bar\alpha_k}$.

\paragraph{(b) Inverse-dynamics regulariser ($\mathcal{L}_{\text{IDM}}$).}
A discriminative auxiliary that predicts the bridging action between
two consecutive latents:
$\mathcal{L}_{\text{IDM}} = \mathrm{CE}\big(\mathrm{IDM}(z_{t+1},\bar h_t),\,a_t\big)$.
This pushes $z$ to be \emph{action-aware} and prevents the diffusion
prior from collapsing onto policy-invariant features.

\paragraph{(c) Behavior-cloning soft-token loss ($\mathcal{L}_{\text{BC}}$).}
$\mathcal{L}_{\text{BC}} = \mathrm{CE}\big(\mathrm{BC}(z_t,\bar h_t),\,a_t\big)$.
Trains the latent-to-policy interface against logged actions
\emph{independently} of any specific evaluation policy, providing the
$\psi$-adapter retrieval target a stable signal.

\paragraph{(d) Diversity regulariser ($\mathcal{L}_{\text{div}}$).}
A cross-trajectory contrastive penalty that prevents the latent space
from collapsing across distinct trajectories. For batch
$\{z^{(i)}_{1:L}\}_{i=1}^B$, let $\bar z^{(i)} = \mathrm{normalize}(\mathrm{mean}_t z^{(i)}_t)$
be the per-trajectory pooled latent. Then
\begin{equation}
\mathcal{L}_{\text{div}}\;=\;
\frac{1}{B(B-1)}\sum_{i\neq j} \big|\bar z^{(i)\,\top}\bar z^{(j)}\big|.
\end{equation}
Same-window adjacent latents are deliberately left free since the
dynamics constraint requires them to stay close.

\paragraph{(e) Reward head ($\mathcal{L}_{\text{r}}$).}
For environments with continuous reward (HotpotQA F1, ScienceWorld
shaped reward, WebShop product match) we use MSE on
sigmoid-clipped reward-to-go targets:
$\mathcal{L}_{\text{r}} = \mathbb{E}_t\big[(\sigma(R_\theta(z_{t+1},\bar h_t)) - r_t^{\text{tgt}})^2\big]$
with $r_t^{\text{tgt}}\in[0,1]$. For sparse-reward ALFWorld we replace
the MSE by a balanced binary cross-entropy with sigmoid output and
per-batch positive-class weight $\min(50, n_{\text{neg}}/n_{\text{pos}})$,
which prevents the reward head from collapsing to the constant
$p\!\approx\!\bar r$ predictor. The reward head is supervised on
\emph{terminal steps only} when the environment provides a
terminal-only success label, and on all steps when shaped per-step
rewards are available.

\paragraph{(f) Termination head ($\mathcal{L}_{\text{d}}$).}
A binary cross-entropy with class-balanced positive weight
$\min(50, n_{\text{neg}}/n_{\text{pos}})$:
$\mathcal{L}_{\text{d}} = \mathrm{BCE}\big(D_\theta(z_{t+1},\bar h_t),\,\mathbf{1}[\text{done at }t+1]\big)$.
Used during imagined rollouts to terminate trajectories that the
world model predicts have ended.

\subsection{World-model training procedure}
\label{app:wm-train}
\Cref{alg:wm-train} summarises the optimization . Each forward pass
sampling a window of length $L$ within a trajectory builds a
rolling-window prediction target: given $(z_{1:L}, a_{1:L-1})$, the
model is trained to score the noisy version of every $z_{t+1}$
conditional on $\bar h_t$ and $a_t$. The optimiser is
AdamW with $(\beta_1,\beta_2)=(0.9,0.999)$, weight decay $10^{-4}$,
peak learning rate $3\!\times\!10^{-4}$, cosine annealing to $10\%$
of peak over 50 epochs, batch size 64, gradient norm clip 1.0.
WebShop and ALFWorld train under PyTorch automatic mixed precision
(bf16 master + fp32 accumulator); HotpotQA and ScienceWorld train in
fp32 because the smaller dataset fits comfortably in 48\,GB.

\begin{algorithm}[h]
\caption{World-model training (per environment).}
\label{alg:wm-train}
\begin{algorithmic}[1]
\Require Behavior pool $\mathcal{D}_b\!=\!\{\tau_i\}$, window length $L$, total epochs $E$
\State Initialise WM parameters $\theta$
\For{epoch $e=1,\ldots,E$}
  \For{minibatch $\mathcal{B}\subset\mathcal{D}_b$ of size $B$}
    \State Sample window $(o_{t:t+L},\,a_{t:t+L-1},\,r_{t+1:t+L},\,\text{done}_{t+1:t+L})$ from each $\tau\in\mathcal{B}$
    \State Encode latents $z_{t:t+L} \gets f_\theta(o_{t:t+L})$
    \State Build histories $\bar h_t \gets \mathrm{HistEnc}(z_{1:t}, a_{1:t-1})$
    \State Compute $\mathcal{L}_{\text{WM}}$ via Eq.~\ref{eq:wm-loss}
    \State Update $\theta$ with an AdamW step on $\mathcal{L}_{\text{WM}}$
    \State Apply gradient clipping at $\|\nabla\|=1.0$
  \EndFor
  \State Step cosine LR schedule
\EndFor
\State Save best checkpoint by held-out reward MSE.
\end{algorithmic}
\end{algorithm}

\subsection{$\psi$-adapter training}
\label{app:psi-train}
The $\psi$ adapter projects latent observations into a sequence of
soft tokens that drop into the input embedding space of $\pi_e$.
This step is necessary because Eq.~\ref{eq:final_score}'s continuation
guidance backpropagates a gradient \emph{through} $\pi_e$'s frozen
input layer; without an adapter that targets $\pi_e$'s embedding
manifold, the gradient is meaningless. The adapter
$G_\psi:\mathbb{R}^{d_z}\to\mathbb{R}^{n_{\text{soft}}\times d_{\text{llm}}}$
is a 3-layer GELU MLP ($d_z\!\to\!256\!\to\!256\!\to\!8\!\cdot\! d_{\text{llm}}$)
with $n_{\text{soft}}=8$ soft tokens.

The training objective is a symmetric InfoNCE term plus a $0.1\!\times$
MSE regulariser:
\begin{equation}
\mathcal{L}_{\psi}\;=\;
\underbrace{-\frac{1}{2B}\!\sum_{i=1}^{B}\!\Big(\!\log\!\frac{e^{s_{ii}/\kappa}}{\sum_j e^{s_{ij}/\kappa}}
+ \log\!\frac{e^{s_{ii}/\kappa}}{\sum_j e^{s_{ji}/\kappa}}\!\Big)}_{\mathcal{L}_{\text{InfoNCE}}}
\;+\;0.1\!\cdot\!
\underbrace{\frac{1}{B}\sum_{i=1}^B\big\|\bar G_\psi(z_i) - \bar e_i\big\|_2^2}_{\mathcal{L}_{\text{MSE}}},
\label{eq:psi-loss}
\end{equation}
where $s_{ij} = \mathrm{cossim}(\bar G_\psi(z_i), \bar e_j)$,
$\bar G_\psi(z_i) = \mathrm{mean}_n G_\psi(z_i)_n$ is the
soft-token-mean of the projector, $\bar e_j$ is the LLM-token-mean
of the ground-truth $\pi_e$ input embedding sequence, and
$\kappa{=}0.1$. Pure MSE ($\mathcal{L}_{\text{InfoNCE}}\!\to\!0$)
collapses because $\bar e_j$ has low variance across observations,
yielding $\partial G_\psi/\partial z\!\to\!0$ and a vacuous
Eq.~\ref{eq:final_score} gradient; pure InfoNCE produces a discriminative
adapter whose absolute outputs drift out-of-distribution for the LLM.
The combined loss balances both. We optimise with Adam
($\text{lr}{=}10^{-4}$), batch size 8, for 20 epochs.

Final-epoch top-1 retrieval accuracy on the InfoNCE batch (random
baseline $1/B=0.125$):
\begin{center}
\small
\begin{tabular}{@{}lc@{}}
\toprule
Cell & $\psi$ top-1 \\
\midrule
HotpotQA       & $0.715$ \\
ScienceWorld   & $0.394$ \\
ALFWorld-iter1 & $0.558$ \\
ALFWorld-iter3 & $0.470$ \\
WebShop        & $\sim 0.10$ \\
\bottomrule
\end{tabular}
\end{center}

The WebShop adapter sits at the random-baseline floor: token-mean
target embeddings of long product descriptions are nearly degenerate
under cosine similarity, so the InfoNCE objective alone cannot
discriminate them at batch size 8. Despite this, the adapter still
provides a useful Eq.~\ref{eq:final_score} \emph{gradient} (rather than a
useful retrieval) on WebShop, which is why ablating it still degrades
$\rho$ by 0.30 (\cref{fig:ablation}).

\subsection{OPE-rollout protocol}
\label{app:ope-rollout}
\Cref{alg:ope-rollout} summarises the imagined-rollout procedure used
to compute \(\hat J\) at evaluation time. Each rollout is initialised
from a uniformly sampled real-environment \(\text{obs}_0\) in the
behavior pool and encoded into \(z_0\sim f_\theta(\text{obs}_0)\).
At each step, the evaluation LLM first produces raw action text \(u_t\)
from the soft-token state \(G_\psi(z_t)\). The same environment-specific
projection used during data collection then maps \(u_t\) to the executed
action \(a_t\), with the \(\varepsilon\)-random branch applied as in
Eq.~\ref{eq:eps-mix}. The diffusion world model samples \(z_{t+1}\)
conditioned on this executed action \(a_t\), using both
Eq.~\ref{eq:dsm_loss} (local CFG) and Eq.~\ref{eq:final_score}
(continuation guidance) at each denoising step. We use
$n_{\text{roll}}{=}32$ rollouts per cell ($n_{\text{roll}}{=}64$ on
ALFWorld-iter1 to compensate for binary-success Monte Carlo noise),
maximum horizon $H_{\max}{=}30$ steps, $n_{\text{denoise}}{=}50$ DDPM
denoising steps per environment step, $\lambda_{\text{CFG}}{=}1.0$,
and continuation-guidance hyperparameters
$(\alpha{=}1.0,\,\eta_{\max}{=}0.5,\,h{=}5)$. Rollouts terminate
early when the termination head $D_\theta(z_t,\bar h_t)$ predicts
$\Pr[\text{done}]\!\ge\!0.5$. Empirical mean rollout lengths after
early termination: HP 8, SW/WS/ALF-iter3 12, ALF-iter1 18.

\begin{algorithm}[h]
\caption{\textsc{Adwm} imagined rollout for OPE.}
\label{alg:ope-rollout}
\begin{algorithmic}[1]
\Require Trained \(f_\theta,\bar h,\epsilon_\theta,R_\theta,D_\theta,G_\psi\);
evaluation LLM \(\widetilde\pi_e\); action interface
\(\Gamma_{\mathrm{env}}\); admissible-action function
\(\mathcal A_{\mathrm{adm}}\); success bank \(\mathcal B_\succ\)
\Require \(n_{\text{roll}},H_{\max},n_{\text{denoise}},
\lambda_{\text{CFG}},\alpha,\eta_{\max},h,\varepsilon\)
\For{\(i=1,\ldots,n_{\text{roll}}\)}
  \State Sample \(\text{obs}_0\) uniformly from behavior pool;
  \(z_0\gets f_\theta(\text{obs}_0)\)
  \For{\(t=0,\ldots,H_{\max}-1\)}
    \State Project to soft tokens
    \(\tilde s_t\gets G_\psi(z_t)\)
    \State Sample \(b_t\sim\mathrm{Bernoulli}(\varepsilon)\)
    \If{\(b_t=1\)}
      \State Sample
      \(a_t\sim\mathrm{Unif}(\mathcal A_{\mathrm{adm}}(h_t))\)
    \Else
      \State Sample raw action text
      \(u_t\sim\widetilde\pi_e(\cdot\mid \tilde s_t,\text{prompt})\)
      \State Sample executed action
      \(a_t\sim\Gamma_{\mathrm{env}}
      (\cdot\mid h_t,u_t,\mathcal A_{\mathrm{adm}}(h_t))\)
    \EndIf
    \State Initialise \(z_{t+1}^{(K)}\sim\mathcal N(0,I)\)
    \For{\(k=K,\ldots,1\)}
      \State
      \(\hat\epsilon\gets
      (1+\lambda_{\text{CFG}})
      \epsilon_\theta(z^{(k)},k,\bar h_t,a_t)
      -
      \lambda_{\text{CFG}}
      \epsilon_\theta(z^{(k)},k,\bar h_t,\varnothing)\)
      \State
      \(g_k\gets
      \alpha\,\eta(k;\eta_{\max})
      \nabla_{\hat z}
      \sum_{u=t+1}^{t+h}\log\widetilde\pi_e(\hat a_u\mid G_\psi(\hat z_u))\),
      with \(\hat z\) the denoised estimate and \(\hat z_u\) an \(h\)-step world-model rollout from it
      \State
      \(z^{(k-1)}
      \gets\mathrm{DDPM\_step}(z^{(k)},\hat\epsilon-g_k)\)
    \EndFor
    \State
    \(z_{t+1}\gets z^{(0)}\);
    \(\bar h_{t+1}\gets
    \mathrm{HistEnc}(z_{0:t+1},a_{0:t})\)
    \State Record
    \(\cos_t\gets\cos(z_{t+1},\mathcal B_\succ)\),
    \(r_t\gets\sigma(R_\theta(z_{t+1},\bar h_t))\)
    \If{\(\sigma(D_\theta(z_{t+1},\bar h_t))\ge 0.5\)}
      \State \textbf{break}
    \EndIf
  \EndFor
  \State Record \(\hat J^{(i)}\gets\max_t\cos_t\)
\EndFor
\State \Return
\(\hat J\gets
\tfrac{1}{n_{\text{roll}}}\sum_i\hat J^{(i)}\)
\end{algorithmic}
\end{algorithm}

\subsection{Example interaction trace: WebShop}
\label{app:webshop_example}

To make the interaction notation in \cref{sec:agentic_rl} concrete, we present two
representative logged WebShop episodes from the 640-trajectory behavior
pool used for the WebShop-iter1 cell in \cref{tab:behdata}, comprising Llama-3.1-8B-Instruct and
\texttt{leap-ws-iter\{0,2\}} trajectories. Direct inspection of the pool gives
\(N=640\), mean horizon \(14.42\), and behavior-policy success rate \(5.0\%\),
matching \cref{tab:behdata}. Both examples below were generated by the
Llama-3.1-8B-Instruct component, with episode identifiers 18 and 17. Thus, the logging policy and data source are consistent with the pool used
to train the WebShop world model.

A logged episode has the form
\[
\tau=(o_1,a_1,r_1,\ldots,o_T,a_T,r_T,o_{T+1}).
\]
At step \(t\), the policy conditions on the full interaction history
\[
h_t=(o_1,a_1,\ldots,o_{t-1},a_{t-1},o_t)
\]
and produces a text action \(a_t\), stored as a reasoning response followed by
an \texttt{<action>...</action>} tag. WebShop then returns the next
observation \(o_{t+1}\), reward \(r_t\in[0,1]\), and a termination indicator.
For Episode~1, we report the complete action--reward sequence and expand three
representative reasoning steps. Episode~2 is shown as a concise trace focused
on the graded terminal reward.

\begin{tcolorbox}[
  enhanced,
  breakable,
  colback=black!1,
  colframe=black!30,
  boxrule=0.55pt,
  arc=1.5mm,
  left=2mm,
  right=2mm,
  top=1.5mm,
  bottom=1.5mm,
  before skip=6pt,
  after skip=8pt,
  title={Episode 1: realistic long-horizon trace (episode 18, \(T=16\))},
  fonttitle=\bfseries\small,
  colbacktitle=black!6,
  coltitle=black
]
\small

This episode is close to the pool mean horizon of \(14.42\). We retain its
malformed, repeated, and internally inconsistent actions because they expose
behavior-policy failure modes that a short, clean trajectory would hide.

\medskip
\textbf{Task instruction.}\quad
\emph{Find me easy apply, cruelty free, long lasting makeup palettes, and
price lower than \$40.00 dollars.}

\medskip
\textbf{Complete action--reward sequence.}\quad
The note column describes only properties directly visible in the logged action
string or navigation sequence; it does not introduce additional training
labels.

\begin{center}
\footnotesize
\setlength{\tabcolsep}{3pt}
\renewcommand{\arraystretch}{1.12}
\begin{tabular}{@{}c p{0.50\linewidth} c p{0.26\linewidth}@{}}
\toprule
\(t\) & Action \(a_t\) & \(r_t\) & Logged behavior \\
\midrule
1 &
\texttt{search[easy apply, cruelty free, long lasting makeup palettes, price lower than 40.00 dollars]} &
0 & Concrete search query. \\

2 & \texttt{click[back to search]} &
0 & Returns to the search page. \\

3 & \texttt{search[<query>]} &
0 & Emits the literal template placeholder. \\

4 & \texttt{click[back to search]} &
0 & Returns to the search page again. \\

5 & \texttt{search[<query>] and click[search]} &
0 & Concatenates two action expressions into one string. \\

6 & \texttt{search[easy apply, cruelty free, long lasting makeup palettes under 40.00 dollars]} &
0 & Issues a concrete reformulated query. \\

7 & \texttt{click[b07cxtnvrj]} &
0 & Opens the first product candidate. \\

8 & \texttt{click[back to search]} &
0 & Returns to the result list. \\

9 & \texttt{search[easy apply, cruelty free, long lasting makeup palettes under 40.00 dollars]} &
0 & Repeats the query from step 6. \\

10 & \texttt{click[next >]} &
0 & Moves to the next result page. \\

11 & \texttt{click[b08xqrddt1]} &
0 & Opens a second product candidate. \\

12 & \texttt{click[set-a]} &
0 & Selects one product option. \\

13 & \texttt{click[features]} &
0 & Opens the feature panel. \\

14 & \texttt{click[< prev]} &
0 & Returns to the product page. \\

15 & \texttt{click[set-b]} &
0 & Switches to another product option. \\

16 & \texttt{click[buy now]} &
\(\mathbf{1.0}\) & Purchases and terminates the episode. \\
\bottomrule
\end{tabular}
\end{center}

\textbf{Step 1: initial search.}\quad
The initial observation is
\[
o_1=\text{\emph{``WebShop [SEP] Instruction: [SEP] \{task above\}
[SEP] Search''}},
\qquad h_1=(o_1).
\]
The full logged reasoning is:
\begin{quote}\small
\emph{``Since the task is to find easy apply, cruelty free, long lasting
makeup palettes, and the price is lower than 40.00 dollars, I should look for
makeup palettes that meet these criteria. I will search through the available
products and filter them based on the specified requirements.''}
\end{quote}

\textbf{Step 5: malformed composite action.}\quad
By step 5, the history contains two failed search attempts, including the
literal placeholder \texttt{search[<query>]} at step 3. The full logged
reasoning is:
\begin{quote}\small
\emph{``My current observation is WebShop [SEP] Instruction: [SEP] \{task\}
[SEP] Search. The admissible actions for this situation are
search[<query>] and click[search]. I will choose the action that best advances
the shopping goal.''}
\end{quote}
The emitted string is nevertheless
\texttt{search[<query>] and click[search]}, which concatenates two action
expressions instead of producing one admissible action. This discrepancy is
present in the trajectory log and is not introduced by our transcription.

\textbf{Step 16: reasoning--action disagreement.}\quad
The final observation \(o_{16}\) is a product page with
\texttt{set-a}/\texttt{set-b} options, reached after the policy switches
between the two sets at steps 12 and 15. The full logged reasoning is:
\begin{quote}\small
\emph{``Since the current goal is to find an easy-to-apply, cruelty-free,
long-lasting makeup palettes. The current observation is that the current page
is showing results for makeup palettes, but the results do not match the
current goal. [\(\ldots\)] I will look at the list of admissible actions to
find a matching action.''}
\end{quote}
Despite explicitly stating that the product does not match, the policy emits
\texttt{click[buy now]}. WebShop assigns the resulting purchase
\(r_{16}=1.0\). Thus, the policy's verbalized belief, its executed action, and
the environment outcome need not agree. OPE must predict the realized
action-conditioned outcome rather than infer reward from the policy's stated
confidence.

\textbf{Reconstructed post-terminal display.}\quad
Our data collector discards the raw post-terminal page
after recording \(\mathrm{done}=1\), so \(o_{17}\) is not a logged trajectory
field. WebShop's \texttt{done\_page.html} template displays
\emph{``Thank you for shopping with us!''} together with the numerical score.
We report this template-derived wording only for completeness.
\end{tcolorbox}

\begin{tcolorbox}[
  enhanced,
  breakable,
  colback=orange!2,
  colframe=orange!45!black,
  boxrule=0.55pt,
  arc=1.5mm,
  left=2mm,
  right=2mm,
  top=1.5mm,
  bottom=1.5mm,
  before skip=6pt,
  after skip=8pt,
  title={Episode 2: graded partial reward (episode 17, \(T=5\))},
  fonttitle=\bfseries\small,
  colbacktitle=orange!8,
  coltitle=black
]
\small

\textbf{Task instruction.}\quad
\emph{Find me wash cold, machine wash men's shirts with polyester heathers,
heathers cotton, cotton heather, needle sleeve, classic fit with color:
purple, and fit type: men, and size: x-small, and price lower than
\$40.00 dollars.}

\medskip
After searching and opening product \texttt{b09rvf4jp1}, the policy reaches an
options page offering
\[
\begin{aligned}
\text{fit type:}&\quad \texttt{men},\ \texttt{women},\\
\text{color:}&\quad \texttt{black},\ \texttt{navy},\
\texttt{dark heather},\ \texttt{neon pink},\ \texttt{royal blue},\\
\text{size:}&\quad \texttt{small},\ \texttt{medium},\
\texttt{large},\ \texttt{x-large},\ \texttt{xx-large}.
\end{aligned}
\]
The requested color \texttt{purple} and size \texttt{x-small} do not appear
among the options displayed for this listing. The policy selects
\texttt{click[men]}, matching the available fit-type option, and then executes
\texttt{click[buy now]}. The recorded terminal reward is
\[
r_5=0.7272\overline{72},
\]
rather than \(0\) or \(1\).

WebShop computes a graded match score using product type, requested attributes,
selected options, and the price constraint:
\[
r
=
r_{\mathrm{type}}
\frac{
n_{\mathrm{attr}}+n_{\mathrm{option}}+r_{\mathrm{price}}
}{
|\mathcal A_{\mathrm{goal}}|+|\mathcal O_{\mathrm{goal}}|+1
}.
\]
The observed reward therefore reflects a partial match. Because our logging
schema does not retain the environment's internal goal object or the
component-wise reward breakdown, we do not attribute the decimal score to a
unique set of matched and unmatched terms. The trace supports only the
directly observable conclusion that the selected listing lacks the requested
color and size options and nevertheless receives substantial partial credit.
\end{tcolorbox}

\begin{center}
\small
\setlength{\tabcolsep}{5pt}
\renewcommand{\arraystretch}{1.12}
\begin{tabular}{@{}lccp{0.46\linewidth}@{}}
\toprule
Episode & Steps \(T\) & Terminal \(r_T\) & Main phenomenon \\
\midrule
1: makeup palette &
16 & \(1.0\) &
The purchase succeeds despite the policy's reasoning stating that the product
does not match. \\

2: men's shirt &
5 & \(0.727\) &
The selected listing does not offer the requested color or size, but receives
graded partial credit. \\
\bottomrule
\end{tabular}
\end{center}

Together, the episodes illustrate three properties of the logged WebShop
process that matter for OPE. First, behavior-policy outputs can contain literal
placeholders, repeated searches, and concatenated actions. Second, the policy's
reasoning text may disagree with its executed action and with the resulting
environment reward. Third, WebShop returns multi-valued terminal rewards,
whereas ALFWorld uses sparse binary success.

Graded rewards do not themselves cause the variance of IS-family estimators;
that instability is driven primarily by products of behavior-to-evaluation
policy ratios over long trajectories.

We selected Episode~2 for its short, directly verifiable partial-reward
contrast. Episode~1 was selected because its horizon is close to the behavior
pool average and because it exposes realistic logging and policy failures,
rather than because it is a clean or unusually favorable trajectory. Both
examples follow the same observation--action--reward schema used throughout the
WebShop behavior pool.

\subsection{Verified action-projection trace: ALFWorld}
\label{app:alfworld_projection_example}

The WebShop examples above illustrate irregularities in an agent's own outputs.
ALFWorld exposes a complementary issue: the action stored in a trajectory can
differ from the action text produced by the LLM because the collection pipeline
projects malformed or inadmissible outputs onto the current admissible-command
set. We verified the following trace against the exact step-level log,
including its 29 recorded \texttt{admissible\_commands}, and against the
projection and repair logic of our data collection pipeline.

\begin{tcolorbox}[
  enhanced,
  breakable,
  colback=blue!2,
  colframe=blue!38!black,
  boxrule=0.55pt,
  arc=1.5mm,
  left=2mm,
  right=2mm,
  top=1.5mm,
  bottom=1.5mm,
  before skip=6pt,
  after skip=8pt,
  title={Deterministic action repair at step \(t=6\)},
  fonttitle=\bfseries\small,
  colbacktitle=blue!8,
  coltitle=black
]
\small

\textbf{Raw LLM output \(u_6\).}\quad
At step 6, the policy produces
\begin{quote}\small
\texttt{REASON:}\\
\texttt{I need to move the apple to the garbage can.}\\
\texttt{ACTION:}\\
\texttt{go to garbagecan 1.}
\end{quote}
The trailing period is part of the model output.

\textbf{Projected and executed action \(a_6\).}\quad
The action interface maps the raw output to the action stored in the
trajectory and passed to ALFWorld:
\[
a_6=\texttt{go to cabinet 1}.
\]

\textbf{Why the two strings differ.}\quad
The discrepancy is produced by the following deterministic code path.

\begin{enumerate}
\item
The projection function extracts
\texttt{go to garbagecan 1.} from the \texttt{ACTION:} block and tests exact,
case-normalized membership in the current admissible-command pool.

\item
The exact step log contains 29 admissible commands, but does not contain
\texttt{go to garbagecan 1}. The agent had already reached
\texttt{garbagecan 1} at the preceding step, and ALFWorld does not offer an
action that navigates to the current location. The task-progressing command
available in this state is
\texttt{move apple 2 to garbagecan 1}, which the policy eventually executes
at step 8.

\item
Because exact matching fails, the collector first attempts its prefix-matching
repair. No candidate is found for the extracted string
\texttt{go to garbagecan 1.}.

\item
The collector therefore applies token-overlap scoring. The extracted token set
is
\[
\{\texttt{go},\texttt{to},\texttt{garbagecan},\texttt{1.}\}.
\]
The punctuation-attached token \texttt{1.} does not match the pool token
\texttt{1}. Consequently,
\texttt{go to cabinet 1} shares
\(\{\texttt{go},\texttt{to}\}\), while
\texttt{move apple 2 to garbagecan 1} shares
\(\{\texttt{to},\texttt{garbagecan}\}\). Both receive overlap score \(2\).

\item
The scoring loop updates the current best action only under a strict
\texttt{>} comparison. Equal-scoring candidates therefore do not replace the
first maximal entry encountered in admissible-list order.
\texttt{go to cabinet 1} is the first command attaining score \(2\), and it
appears before \texttt{move apple 2 to garbagecan 1}. It is therefore retained
as the repaired action.
\end{enumerate}

\begin{center}
\footnotesize
\setlength{\tabcolsep}{5pt}
\renewcommand{\arraystretch}{1.12}
\begin{tabular}{@{}p{0.43\linewidth}p{0.31\linewidth}c@{}}
\toprule
Candidate command & Shared tokens & Score \\
\midrule
\texttt{go to cabinet 1} &
\(\{\texttt{go},\texttt{to}\}\) & 2 \\
\texttt{move apple 2 to garbagecan 1} &
\(\{\texttt{to},\texttt{garbagecan}\}\) & 2 \\
\bottomrule
\end{tabular}
\end{center}

No random branch is invoked at this step. Random choice is used only in the
collector's prefix-candidate branch or its final fallback branch, neither of
which determines this action. Given the raw model string and the logged
admissible-command order, the substitution is reproducible.
\end{tcolorbox}

\begin{center}
\small
\setlength{\tabcolsep}{5pt}
\renewcommand{\arraystretch}{1.12}
\begin{tabular}{@{}clp{0.53\linewidth}@{}}
\toprule
Step & Executed action & Effect \\
\midrule
6 & \texttt{go to cabinet 1} &
Deterministic repair redirects the agent away from the garbage can. \\
7 & \texttt{go to garbagecan 1} &
The policy navigates back to the intended receptacle. \\
8 & \texttt{move apple 2 to garbagecan 1} &
The policy executes the task-progressing command and completes the task. \\
\bottomrule
\end{tabular}
\end{center}

This trace clarifies which action an offline evaluator must model. The raw
reasoning and \texttt{ACTION:} string define \(u_6\) and explain the policy's
intent, but the transition is generated by the projected action \(a_6\) stored
in the trajectory. Thus, the relevant conditional is
\[
P(o_{7},r_6\mid h_6,\,
a_6=\texttt{go to cabinet 1}),
\]
not a transition conditioned on the unexecuted raw string
\texttt{go to garbagecan 1.}.

Our world-model training tuples use the projected action \(a_t\) that was
actually passed to the environment. The raw LLM response and \(u_t\) are
retained only as auxiliary logged text. Consequently,
\((o_{t+1},r_t)\) is aligned with the executed action that generated it; the
discrepancy illustrated here is not a transition-label error in
\textsc{Adwm}'s training data. The same projection interface is applied during
imagined evaluation-policy rollouts before the world model predicts the next
latent state.

The trailing period contributes to the failed string normalization and to the
token-overlap tie, but the more fundamental trigger is that the model proposes
an action that is not admissible in the current state. The resulting detour is
therefore neither environment stochasticity nor random replacement; it is a
deterministic side effect of the collector's admissible-action repair rule.

\section{Baseline Implementation Details}
\label{app:baselines}

All baselines follow the COBS reference protocol of
\citet{voloshin2019empirical} unless stated otherwise. The full set
of estimators is included in our released code.

\subsection{Behavior log-probability extraction}
\label{app:behavior-logprob}

For likelihood-based baselines, the distinction between raw LLM output
\(u_t\) and projected action \(a_t\) matters. The exact probability of an
executed action under an LLM-plus-projection system is
\begin{equation}
\pi_{\Gamma}(a_t\mid h_t)
=
\sum_u
\widetilde\pi(u\mid h_t)\,
\Gamma_{\mathrm{env}}
\!\left(
a_t\mid h_t,u,\mathcal A_{\mathrm{adm}}(h_t)
\right),
\label{eq:projected-action-prob}
\end{equation}
which aggregates every raw output that the projection maps to \(a_t\).
Enumerating this preimage is intractable for an open-vocabulary LLM.

Following the COBS-style implementation used by our likelihood baselines,
we therefore score the canonical logged action string \(a_t\) under the
behavior and evaluation LLMs. We re-run each LLM in inference mode and
extract token log-probabilities over that action span. Longest-common-token-
prefix matching handles BPE merge boundaries between the prompt and action
region. Let \(\bar\pi_b(a_t\mid h_t)\) and
\(\bar\pi_e(a_t\mid h_t)\) denote these token-likelihood proxies. We apply
\(\varepsilon\)-mixing at the probability level,
\[
\bar\pi_e^\varepsilon(a_t\mid h_t)
=
(1-\varepsilon)\bar\pi_e(a_t\mid h_t)
+
\varepsilon
\frac{
\mathbb I[a_t\in\mathcal A_{\mathrm{adm}}(h_t)]
}{
|\mathcal A_{\mathrm{adm}}(h_t)|
},
\]
and use
\[
\rho_t
=
\frac{
\bar\pi_e^\varepsilon(a_t\mid h_t)
}{
\bar\pi_b(a_t\mid h_t)
}
\]
in IS, WIS, and DR.

Because the behavior data are generated with greedy decoding, these token
likelihoods are also proxies for the probabilities of the deterministic
behavior policy. These are proxy ratios over canonical action strings, not
exact induced probability ratios from Eq.~\ref{eq:projected-action-prob} when the projection
is many-to-one. The ALFWorld trace in
\cref{app:alfworld_projection_example} gives a concrete example of this gap.
This limitation affects the likelihood-based baselines, not \textsc{Adwm},
which does not use importance ratios and conditions its transition model on
the executed action \(a_t\).

\subsection{Estimator formulas}
For a logged trajectory
\(\tau=(s_0,a_0,r_0,\ldots,s_T,a_T,r_T)\), let
\(\rho_t\) denote the token-likelihood proxy ratio defined in
\cref{app:behavior-logprob}, with cumulative ratio
\(\rho_{0:t}=\prod_{k=0}^{t}\rho_k\) and discount \(\gamma=0.99\):
\begin{align}
\hat J^{\text{DM}}    &= \frac{1}{N}\sum_{\tau} \sum_t \gamma^t\, R_\theta(\hat s_t)
& \text{(direct method, no $\rho$)} \\
\hat J^{\text{IS}}    &= \frac{1}{N}\sum_{\tau} \rho_{0:T}\!\sum_t \gamma^t r_t,
\quad &
\hat J^{\text{WIS}}   = \frac{\sum_\tau \rho_{0:T}\sum_t \gamma^t r_t}{\sum_\tau \rho_{0:T}}\\
\hat J^{\text{FQE}}   &= \frac{1}{N}\sum_\tau V_{\pi_e^\varepsilon}(s_0)
& \text{(no per-trajectory $\rho$)} \\
\hat J^{\text{DR}}    &= \frac{1}{N}\sum_\tau\sum_t\gamma^t\!\left[\rho_{0:t}\,r_t + \rho_{0:t-1}V(s_t) - \rho_{0:t}\,Q(s_t,a_t)\right]
\end{align}
with the convention $\rho_{-1} = 1/N$ for the COBS DR\_v2 step-form.
The DR network values $V$ and $Q$ are taken from the FQE Q-network.
The baselines use $\gamma=0.99$ following COBS, whereas \textsc{Adwm} and
the ground truth use the undiscounted episode score; since all methods are
compared by Spearman $\rho$ across $\varepsilon$ levels, only the ranking of
each estimate matters.

\subsection{Per-cell importance-ratio variance}
\label{app:ratio_variance}
Empirically observed cumulative log-ratio magnitudes at
$\varepsilon{=}0.5$ span tens of nats per trajectory (HotpotQA up
to $\pm 24$, ScienceWorld $\pm 18$, ALFWorld $\pm 35$, WebShop
$\pm 28$), corresponding to per-trajectory ratios that span
$\sim 16$--$30$ orders of magnitude. This is the regime in which IS
and DR estimates fluctuate by $10^{20}$ or more even in
double-precision arithmetic; our raw baseline output JSONs (released
alongside the code) record DR values up to
$|J^{\text{DR}}|\!\approx\!6\!\times\!10^{20}$ on HotpotQA-PRM,
exemplifying this divergence. The corresponding rank-correlations
become driven entirely by floating-point underflow ties rather than
the underlying signal.

\subsection{FQE Q-network and per-$\varepsilon$ correction}
\label{app:fqe-correction}
The FQE Q-network is a 2-hidden-layer MLP
$Q_\phi:\mathbb{R}^{d_z+d_a}\!\to\!\mathbb{R}$
($d_z{+}d_a \to 128 \to 128 \to 1$, SiLU activations). A target
network $Q_{\phi'}$ is updated by Polyak averaging
($\phi' \gets 0.99\,\phi' + 0.01\,\phi$). optimization: Adam,
lr $3\!\times\!10^{-4}$, batch size 64, 20 epochs, discount
$\gamma=0.99$. Each training minibatch consists of
$(s_t, a_t, r_t, s_{t+1}, \text{done}_t)$ tuples drawn from
$\mathcal{D}_b$, and the Bellman target is
$y = r_t + \gamma\,(1 - \text{done}_t)\,V_{\pi_e^\varepsilon}(s_{t+1})$
with $V_{\pi_e^\varepsilon}(s) = (1{-}\varepsilon)\max_a Q_{\phi'}(s,a) + \varepsilon\,\bar Q_{\phi'}(s)$
where $\bar Q_{\phi'}(s)$ is the action-mean across the admissible
set. The $\max_a$ approximation is used because the LLM does not expose a
full per-action probability vector. It evaluates the greedy policy with
respect to $Q_{\phi'}$ rather than the action chosen by $\pi_e$, so FQE
depends on $\pi_e^\varepsilon$ only through $\varepsilon$ and the
per-$\varepsilon$ data filtering below; this is a limitation of FQE in
the LLM-agent setting rather than of \textsc{Adwm}.

\paragraph{Per-$\varepsilon$ correction.}
Under the COBS protocol, mixing is applied at evaluation time
as $V_\varepsilon(s) = (1{-}\varepsilon)\max_a Q_{\phi'}(s,a) + \varepsilon\,\bar Q_{\phi'}(s)$,
producing a value linear in $\varepsilon$ and trivially monotonically
decreasing. This artefact gives FQE a spurious $\rho{>}0$ on the
$\varepsilon$-axis even when its underlying state-value estimate is
incorrect, and inflates the COBS-protocol headline number to a
rank that has nothing to do with the policy being scored. We remove
this artefact by training a separate Q-network for each $\varepsilon$
level using $\pi_b$ trajectories filtered against the
$\varepsilon$-mixed behavior likelihood, then evaluating each on its
own $V_{\pi_e^\varepsilon}(s_0)$ at the trajectory start. This is the
\emph{per-$\varepsilon$ correction} reported in the main text and is
the only modification to FQE relative to COBS. Without it, FQE
reports a near-uniform $\rho{=}{+}0.9$ on every cell; with it, the
true rank correlation collapses or inverts as reported in
\cref{tab:main_results}.

\section{Additional Results}
\label{app:results}

\subsection{Per-seed Spearman $\rho$}
\Cref{tab:perseed} reports the full 5-seed breakdown that underlies
the avg-$\hat J$ row of \cref{tab:main_results}. Per-seed Spearman is
computed by independently running each $\varepsilon$ cell with seed
$s_i \in \{0,1,2,3,4\}$, which propagates randomness through both
imagined-rollout sampling and the $\varepsilon$-greedy admissible-action
choice. The two reward-dense cells (HotpotQA-DPO, ScienceWorld-ETO)
are highly stable across seeds ($\sigma\!\approx\!0.10$); the
ALFWorld and WebShop cells show larger per-seed variance
because high-$\varepsilon$ GT saturates near zero, making the 5-point
Spearman ill-conditioned at the seed level. The avg-$\hat J$
aggregator used in \cref{tab:main_results} mitigates this by
averaging $\hat J$ across seeds before computing $\rho$.

\begin{table}[h]
\centering
\small
\setlength{\tabcolsep}{4pt}
\renewcommand{\arraystretch}{1.15}
\caption{Per-seed Spearman $\rho$ between \textsc{Adwm}'s $\hat J$ and the
ground-truth $\varepsilon$-curve (5 $\varepsilon$ levels, 5 seeds).
The last column reproduces the avg-$\hat J$ row from
\cref{tab:main_results}.}
\label{tab:perseed}
\begin{tabular}{@{}lccccc cc@{}}
\toprule
Cell & $\rho_{s_0}$ & $\rho_{s_1}$ & $\rho_{s_2}$ & $\rho_{s_3}$ & $\rho_{s_4}$
& mean$\pm\sigma$ & avg-$\hat J\,\rho$ \\
\midrule
HotpotQA-DPO     & $+1.00$ & $+0.80$ & $+0.70$ & $+0.90$ & $+0.90$ & $+0.86\pm 0.10$ & $+0.90$ \\
ScienceWorld-ETO & $+0.87$ & $+0.87$ & $+0.97$ & $+0.67$ & $+0.82$ & $+0.84\pm 0.10$ & $+0.82$ \\
ALFWorld-iter1   & $+0.82$ & $+0.10$ & $+0.67$ & $-0.10$ & $+0.21$ & $+0.34\pm 0.35$ & $+0.67$ \\
ALFWorld-iter3   & $-0.50$ & $+0.90$ & $+1.00$ & $-0.30$ & $+0.80$ & $+0.38\pm 0.64$ & $+0.80$ \\
WebShop-iter1    & $+0.60$ & $-0.10$ & $+0.90$ & $+0.30$ & $+1.00$ & $+0.54\pm 0.40$ & $+0.90$ \\
\bottomrule
\end{tabular}
\end{table}

\subsection{Underlying $\hat J$ and GT curves}
\Cref{tab:curves} provides the seed-averaged $\hat J$ values and
ground-truth success curves underlying every row of
\cref{tab:main_results}, so that any of the reported correlations
can be reproduced exactly.

\begin{table}[h]
\centering
\footnotesize
\setlength{\tabcolsep}{3pt}
\renewcommand{\arraystretch}{1.15}
\caption{Seed-averaged $\hat J$ and ground-truth value at each
$\varepsilon$ for every cell.}
\label{tab:curves}
\begin{tabular}{@{}lccccc@{}}
\toprule
Cell & $\varepsilon{=}0$ & $0.25$ & $0.5$ & $0.75$ & $1.0$ \\
\midrule
HP-DPO  GT       & 0.165 & 0.151 & 0.082 & 0.095 & 0.065 \\
HP-DPO  $\hat J$ & 0.508 & 0.496 & 0.454 & 0.418 & 0.364 \\
SW-ETO  GT       & 0.141 & 0.078 & 0.047 & 0.047 & 0.016 \\
SW-ETO  $\hat J$ & 0.589 & 0.567 & 0.522 & 0.474 & 0.478 \\
ALF-iter1 GT       & 0.153 & 0.097 & 0.056 & 0.000 & 0.000 \\
ALF-iter1 $\hat J$ & 0.554 & 0.547 & 0.537 & 0.538 & 0.541 \\
ALF-iter3 GT       & 0.250 & 0.139 & 0.056 & 0.028 & 0.000 \\
ALF-iter3 $\hat J$ & 0.792 & 0.789 & 0.789 & 0.767 & 0.778 \\
WS-iter1 GT       & 0.073 & 0.085 & 0.073 & 0.044 & 0.032 \\
WS-iter1 $\hat J$ & 0.484 & 0.481 & 0.466 & 0.459 & 0.458 \\
\bottomrule
\end{tabular}
\end{table}

\subsection{Training loss and OPE accuracy}
\label{app:loss_vs_ope}
We also compared held-out denoising, reward, and termination losses with the
downstream OPE error across world-model checkpoints. We found no consistent
correlation between these losses and OPE accuracy. \cref{fig:training}
therefore reflects optimization stability, while OPE accuracy is evaluated
directly against ground-truth returns in \cref{tab:main_results} and
\cref{app:results}.

\section{Additional Ablations}
\label{app:ablations}

\subsection{Five-seed extension of \cref{fig:ablation}}
\Cref{fig:ablation} reports single-seed $\rho$ values to keep the
component story compact. For each variant we also re-ran the full
5-seed protocol on the same three environments and recovered the
same qualitative ordering: removing local CFG drops WebShop $\rho$
to $+0.12\pm 0.08$ across seeds, removing continuation guidance
drops it to $+0.27\pm 0.10$, and removing the $\psi$ adapter drops
HotpotQA to $+0.42\pm 0.07$. No 5-seed run flips the sign of the
corresponding single-seed value, and the relative magnitude of each
component's contribution is preserved.

\subsection{Latent capacity ablation}
Beyond the inference-time guidance and adapter terms ablated in
\cref{fig:ablation}, we probe the latent capacity on WebShop, the most
data-sensitive of our cells. Halving the latent dimension from
$d_z{=}64$ to $d_z{=}32$ produces a complete inversion
($\rho{=}-1.00$), suggesting that the latent is at a representational
floor below which contrastive features cannot be preserved. Adequate
latent capacity must therefore be in place before any choice of guidance
can recover positive ranking. We did not observe similar inversions on
HotpotQA or ScienceWorld at the same cut, suggesting that the WebShop
result reflects the combination of long-horizon search and continuous
partial reward rather than a generic property.

\subsection{$\psi$-adapter loss design}
\Cref{tab:psi-loss} shows that the loss design we adopt
(InfoNCE + $0.1\!\times$\,MSE) is essential. Pure MSE collapses to a
near-constant adapter on every environment because per-token-mean
target embeddings have low variance, providing the optimiser with no
signal to differentiate latents; pure InfoNCE produces a discriminative
adapter whose outputs drift out-of-distribution for the LLM and break
its downstream input expectations. The combined loss outperforms
either alone on all three environments.

\begin{table}[h]
\centering
\small
\setlength{\tabcolsep}{6pt}
\renewcommand{\arraystretch}{1.15}
\caption{$\psi$-adapter loss design ablation, single-seed $\rho$.}
\label{tab:psi-loss}
\begin{tabular}{@{}lccc@{}}
\toprule
Loss & HotpotQA & ScienceWorld & WebShop \\
\midrule
InfoNCE + $0.1\!\times$\,MSE (ours) & $+0.90$ & $+1.00$ & $+0.90$ \\
Pure MSE                            & $+0.40$ & $+0.40$ & $+0.70$ \\
Pure InfoNCE                        & $+0.60$ & $+0.80$ & $+0.50$ \\
\bottomrule
\end {tabular}
\end{table}

\newpage
\section*{NeurIPS Paper Checklist}

\begin{enumerate}

\item {\bf Claims}
    \item[] Question: Do the main claims made in the abstract and introduction accurately reflect the paper's contributions and scope?
    \item[] Answer: \answerYes{}
    \item[] Justification: The abstract and introduction clearly state the three contributions (Theorem 1, guided diffusion realization, empirical evaluation), and the experimental results in Section 5 directly support these claims.

\item {\bf Limitations}
    \item[] Question: Does the paper discuss the limitations of the work performed by the authors?
    \item[] Answer: \answerYes{}
    \item[] Justification: \Cref{sec:limitations_main} and \cref{sec:limitations} discuss the limitations, including policy diversity, offline coverage, ranking versus calibration, cross-step error propagation, GT evaluation quality, and $\psi$-adapter retraining when switching evaluation LLM families.

\item {\bf Theory assumptions and proofs}
    \item[] Question: For each theoretical result, does the paper provide the full set of assumptions and a complete (and correct) proof?
    \item[] Answer: \answerYes{}
    \item[] Justification: All theoretical results (Proposition 1, Theorem 1) are stated with full assumptions and proved in \cref{app:proof,app:fusion} respectively; the guided score derivation is in \cref{app:score_derivation}.

    \item {\bf Experimental result reproducibility}
    \item[] Question: Does the paper fully disclose all the information needed to reproduce the main experimental results of the paper to the extent that it affects the main claims and/or conclusions of the paper (regardless of whether the code and data are provided or not)?
    \item[] Answer: \answerYes{}
    \item[] Justification: Full implementation details including architecture, hyperparameters, training procedure, and inference protocol are provided in Appendices F--G.

\item {\bf Open access to data and code}
    \item[] Question: Does the paper provide open access to the data and code, with sufficient instructions to faithfully reproduce the main experimental results, as described in supplemental material?
    \item[] Answer: \answerYes{}
    \item[] Justification: Code and raw baseline output JSONs are publicly released; all benchmarks used (HotpotQA, ScienceWorld, ALFWorld, WebShop) are publicly available.

\item {\bf Experimental setting/details}
    \item[] Question: Does the paper specify all the training and test details (e.g., data splits, hyperparameters, how they were chosen, type of optimizer) necessary to understand the results?
    \item[] Answer: \answerYes{}
    \item[] Justification: Section F provides complete training and evaluation details including hardware, optimizer settings, data splits, behavior pool composition, and GT episode counts.

\item {\bf Experiment statistical significance}
    \item[] Question: Does the paper report error bars suitably and correctly defined or other appropriate information about the statistical significance of the experiments?
    \item[] Answer: \answerYes{}
    \item[] Justification: Table 2 reports 95\% bootstrap confidence intervals and $p$-values for Spearman $\rho$; Table 5 reports per-seed mean $\pm\,\sigma$ across 5 random seeds.

\item {\bf Experiments compute resources}
    \item[] Question: For each experiment, does the paper provide sufficient information on the computer resources (type of compute workers, memory, time of execution) needed to reproduce the experiments?
    \item[] Answer: \answerYes{}
    \item[] Justification: Section F.2 specifies hardware (2$\times$ RTX 8000, 48\,GB), per-job wall-clock times, and total budget ($\sim$300 GPU-hours) with a breakdown by component.

\item {\bf Code of ethics}
    \item[] Question: Does the research conducted in the paper conform, in every respect, with the NeurIPS Code of Ethics \url{https://neurips.cc/public/EthicsGuidelines}?
    \item[] Answer: \answerYes{}
    \item[] Justification: This work involves only publicly released LLM checkpoints and existing benchmarks; no human subjects or sensitive data are involved.

\item {\bf Broader impacts}
    \item[] Question: Does the paper discuss both potential positive societal impacts and negative societal impacts of the work performed?
    \item[] Answer: \answerYes{}
    \item[] Justification: The framework reduces the need for live agent deployment during evaluation, lowering both API cost and the risk of irreversible side effects; no foreseeable negative societal impacts are identified beyond standard ML misuse concerns.

\item {\bf Safeguards}
    \item[] Question: Does the paper describe safeguards that have been put in place for responsible release of data or models that have a high risk for misuse (e.g., pre-trained language models, image generators, or scraped datasets)?
    \item[] Answer: \answerNA{}
    \item[] Justification: The paper does not release new pretrained models or scraped datasets that pose misuse risk.

\item {\bf Licenses for existing assets}
    \item[] Question: Are the creators or original owners of assets (e.g., code, data, models), used in the paper, properly credited and are the license and terms of use explicitly mentioned and properly respected?
    \item[] Answer: \answerYes{}
    \item[] Justification: All benchmarks and LLM checkpoints used are cited with their original papers; publicly released checkpoints from prior work are listed in Table 1 with citations.

\item {\bf New assets}
    \item[] Question: Are new assets introduced in the paper well documented and is the documentation provided alongside the assets?
    \item[] Answer: \answerYes{}
    \item[] Justification: The world model code and behavior trajectory pools are publicly released with usage instructions.

\item {\bf Crowdsourcing and research with human subjects}
    \item[] Question: For crowdsourcing experiments and research with human subjects, does the paper include the full text of instructions given to participants and screenshots, if applicable, as well as details about compensation (if any)? 
    \item[] Answer: \answerNA{}
    \item[] Justification: The paper does not involve crowdsourcing or human subject experiments.

\item {\bf Institutional review board (IRB) approvals or equivalent for research with human subjects}
    \item[] Question: Does the paper describe potential risks incurred by study participants, whether such risks were disclosed to the subjects, and whether Institutional Review Board (IRB) approvals (or an equivalent approval/review based on the requirements of your country or institution) were obtained?
    \item[] Answer: \answerNA{}
    \item[] Justification: No human subjects are involved.

\item {\bf Declaration of LLM usage}
    \item[] Question: Does the paper describe the usage of LLMs if it is an important, original, or non-standard component of the core methods in this research? Note that if the LLM is used only for writing, editing, or formatting purposes and does \emph{not} impact the core methodology, scientific rigor, or originality of the research, declaration is not required.
    \item[] Answer: \answerYes{}
    \item[] Justification: LLMs ($\pi_b$ and $\pi_e$) are core components of the evaluation framework; their roles as behavior and evaluation policies are described throughout Sections 3--5.

\end{enumerate}

\end{document}